# Developing Conversational Speech Systems for Robots to Detect Speech Biomarkers of Cognition in People Living with Dementia

Rohith Perumandla[1], Young-Ho Bae[2,3], Diego Izaguirre[1], Esther Hwang[1], Andrew Murphy[1], Long-Jing Hsu[4], Selma Sabanovic[4], Casey C. Bennett[1*]

[1]Department of Computing & Digital Media, DePaul University, Chicago, IL USA

[2]Department of Data Science, Hanyang University, Seoul, Korea

[3]AI Technology Center, POSCO DX, Pohang, Korea

[4]School of Informatics, Computing, & Engineering, Indiana University, Bloomington IN, USA

*Corresponding Author: Casey C. Bennett, [cbenne33@DePaul.edu](mailto:cbenne33@DePaul.edu)

## Abstract

This study presents the development and testing of a conversational speech system designed for robots to detect speech biomarkers indicative of cognitive impairments in people living with dementia (PLwD). The system integrates a backend Python WebSocket server and a central core module with a large language model (LLM) fine-tuned for dementia to process user input and generate robotic conversation responses in real-time in less than 1.5 seconds. The frontend user interface, a Progressive Web App (PWA), displays information and biomarker score graphs on a smartphone in real-time to human users (PLwD, caregivers, clinicians). Six speech biomarkers based on the existing literature - Altered Grammar, Pragmatic Impairments, Anomia, Disrupted Turn-Taking, Slurred Pronunciation, and Prosody Changes - were developed for the robot conversation system using two datasets, one that included conversations of PLwD with a human clinician (DementiaBank dataset) and one that included conversations of PLwD with a robot (Indiana dataset). We also created a composite speech biomarker that combined all six individual biomarkers in a single score. The speech system's performance was first evaluated on the DementiaBank dataset showing moderate correlation with MMSE scores, with the composite biomarker score outperforming individual biomarkers. Analysis of the Indiana dataset revealed higher and more variable biomarker scores, suggesting potential differences due to study populations (e.g. severity of dementia) and the conversational scenario (human-robot conversations are different from human-human). The findings underscore the need for further research on the impact of conversational scenarios on speech biomarkers and the potential clinical applications of robotic speech systems.

# 1. Introduction

Alzheimer's and related dementias (ADRD) represent one of the most significant public health issues in the United States and worldwide, given the rapidly aging populations in most developed countries, with nearly 7 million individuals in the US alone [1,2]. A number of researchers in the past 10 years have begun working on conversational speech systems for people living with dementia (PLwD) and Alzheimer's, within the field of conversational user interfaces (CUI) and beyond [3-7]. Those systems are also sometimes referred to as spoken dialogue systems (SDS) or conversational agents (CA). More recently, people have started attempting to deploy such speech systems onboard robots to interact with PLwD and their caregivers, utilizing large language models (LLMs) to allow those robots to engage more flexibly in conversation about a wide range of topics. However, researchers have also begun to report problems with such LLM-based systems, such as keeping the artificial agent on-topic, appropriate turn-taking (not interrupting speech), integration of multiple modalities (e.g. non-verbal cues), and so forth [8,9]. Indeed, it is becoming increasingly apparent that those "dementia speech systems" will need to be fine-tuned to handle the altered communication styles of PLwD – e.g. prosody changes, slurred pronunciation, disrupted turn-taking (elongated interpausal units [IPUs]) – that get progressively worse as their condition advances over time. Out-of-the-box LLMs are insufficient in that regard. At the same time, since those speech changes in PLwD are thought to be related to nerve cell failure caused by dementia [10, 11], there is an opportunity to develop methods of extracting information about PLwD's cognitive health from the conversations themselves. As such, there remain significant research opportunities going forward in two parallel ways: 1) creating fine-tuned dementia speech systems for robots and 2) extracting cognitive information from the human-robot conversations.

One way we can extract information about cognition from conversations is what are known as "linguistic biomarkers" or "speech biomarkers" in the speech community, which typically have been extracted from PLwD speech during recorded conversations with other people [12,13,14]. Such biomarkers have been identified as potential methods for diagnosing dementia, as well as differential diagnosis of various sub-types of dementia, e.g. Lewy Body dementia [13]. At the same time, there has been much recent research on using social robots for dementia to assist both PLwD and their caregivers via human-robot interaction (HRI) [15]. That opens up exciting opportunities to potentially extract speech biomarkers from real-time conversations between humans and robots in the future, averse to relying only on recorded conversations between humans. Such real-time information could be used to help patients better understand their own health on a daily basis, enhance clinical care, and/or link specific activities a social robot does in user homes to improved cognitive functioning of the PLwD. Importantly, the use of conversational robots for that purpose can allow us to create *replicable* conversations across participants or scenarios [16], which is an important capability for scientific research to advance our understanding of how cognitive health affects speech dysfunction in ADRD and aging in general.

However, there are a number of technical challenges to creating conversational robots and speech biomarkers that can be derived from real-time human-robot conversations. The focus of this paper is on the development of such a system and initial testing of it on multiple speech corpuses, including both human-human conversation and human-robot conversation. We explore different possible types of speech biomarkers, evaluate different methods of calculating such biomarkers in real-time onboard a robot, and provide empirical results of the effectiveness of the system in different conversational scenarios (human-human vs. human-robot).

## 2. Background

In this section, we discuss some of the prior research on speech biomarkers in ADRD in Section 2.1, as well as provide some background regarding research on HRI and robots for dementia in general in Section 2.2.

### 2.1 Speech Biomarkers in ADRD

ADRD is a progressive neurodegenerative condition characterized by the decline of cognitive functions such as memory, reasoning, and communication, significantly affecting daily life and activities of those individuals [17]. ADRD results from damage to nerve cells in the brain in older adults, which impairs cognitive speech functions and the ability to communicate effectively [18]. Early signs of dementia often include memory loss, particularly difficulty recalling recent events or information during conversation, which is one of the most common initial symptoms [19]. Moreover, nerve cell deterioration can also lead to loss of physical control of the vocal cords and mouth muscles, leading to changes in pitch and prosody during speech [10,11]. Other early indicators often include impaired executive functioning, such as challenges in planning, problem-solving and language deficits such as difficulty retrieving words, constructing sentences, and maintaining coherent communication during dialogue with others [20,21].

Speech dysfunction has emerged as one of the earliest and most reliable indicators of cognitive decline in dementia, often occurring even before formal diagnosis [17-19]. Speech is intricately linked to cognitive abilities, and changes in linguistic patterns provide valuable insights into the progression of cognitive decline, making it a critical area of research for understanding and diagnosing dementia. This study focuses on six key speech biomarkers, selected for their relevance in detecting and quantifying speech impairments associated with cognitive decline: **altered grammar**, **pragmatic impairments**, **anomia**, **prosody changes**, **slurred pronunciation**, and **turn-taking difficulties**. Each of these biomarkers captures distinct linguistic or acoustic characteristics, which if combined could theoretically provide a comprehensive approach to analyzing the speech patterns of PLwD. Moreover, such speech analysis could potentially be enabled for real-time extraction of biomarkers of cognitive impairments in PLwD, though consideration must be given to processing times if such speech biomarker system is to deploy in the real-world. We describe our approach to such computational challenges in Section 3.

More broadly, speech biomarkers can be categorized into two types: (1) **Content-based features**, including grammar, syntax, lexical complexity (i.e. vocabulary), and semantic coherence (i.e. pragmatics) and (2) **Acoustic-based features**, such as prosody and slurred pronunciation. Additionally, turn-taking difficulties can be seen as their own category. We provide a review of the background literature for all three categories below, as well as prior research attempts to combine *multiple* biomarkers across categories.

### 2.1.1 Content-based features

Previous research has indicated that content-based (i.e. text-based) linguistic characteristics can effectively differentiate between healthy individuals and those with Alzheimer's disease or other forms of dementia [22]. For instance, analyzing grammatical structures and the frequency of use of specific parts of speech, such as nouns, verbs, and pronouns can reveal patterns associated with dementia [23]. Studies have shown that individuals with dementia may exhibit **Altered Grammar**, such as a decrease in use of complex grammatical structures, fewer unique words, and changed adjective/verb clause proportions [24]. Additionally, reduced lexical diversity and decreased syntactic complexity have been observed in the speech of individuals with dementia [25,26]. Extensive research has been conducted on detecting dementia using grammatical linguistic features and machine learning (ML) algorithms. Such studies have leveraged features such as syntactic complexity, lexical diversity, and specific parts of speech to differentiate individuals with ADRD or Mild Cognitive Impairment (MCI) from healthy controls. Various ML classification models - including Support Vector Machines (SVM), Random Forest Classifiers, and Deep Neural Networks - have been utilized in that research, highlighting the effectiveness of these approaches in distinguishing between these groups based on altered grammar usage [22,27,28].

With regard to Altered Grammar, we used some of the above prior research here to implement a biomarker for Altered Grammar that measures changes in syntactic complexity during conversations with a robot. Details of that are explained in Section 3.

Older adults often produce tangential or off-topic utterances during conversations, deviating from the main topic or failing to maintain coherent focus (i.e. **Pragmatic Impairments**), which represents another potential content-based speech biomarker. This pattern is linked to age-related cognitive decline, particularly in the prefrontal cortex, which affects inhibitory control of speech and other behaviors [29]. These challenges are amplified in individuals with dementia, impacting their ability to organize thoughts, recall relevant information, and maintain coherence in long dialogue [30]. Research by Dijkstra et al. (2004) further supports these findings, showing that individuals with dementia exhibit a higher frequency of discourse-impairing features, such as repetitions and vague expressions, reflecting difficulties in maintaining topic focus and using language effectively, ultimately resulting in a loss of conversational coherence [31].

Research on detecting conversational coherence remains relatively limited. To address this, we adapted the coherence function from Hoffman et al. (2018) [32], originally implemented in R, to measure topic semantic coherence in older adults' conversations about seasons. By integrating this function into Python, we calculated the coherence between user utterances and the speech system as a measure of pragmatic impairment, refining the approach to align with our study objectives. We discuss this more in the Methods section below (Section 3).

**Anomia**, defined as difficulties in recalling words or phrases within one's vocabulary, is a key feature of what is known as semantic dementia (SD) [33]. SD is a neurodegenerative disorder that typically results in fluent progressive aphasia. In addition to SD, anomia is frequently observed as a symptom in various cerebral disorders, including ADRD [34]. It is characterized by an impaired ability to name objects, retrieve specific words, and comprehend word meanings, underscoring its role not only as a language deficit but also as a sign of broader semantic processing challenges [35-37]. Researchers have employed diverse methods to study anomia, such as picture naming tasks and the analysis of spontaneous speech [37]. Furthermore, neuroimaging techniques (e,g, MRI, PET) have been utilized to identify the brain regions involved in different forms of anomia [38]. In terms of clinical presentation, anomia disrupts the natural flow of speech by causing word retrieval difficulties, which manifest as hesitations, pauses, and the use of filler words like "umm," "hmm," and "ah," thereby reducing fluency [35,39]. Those experiencing anomia often use non-specific terms such as "thing" or "place" and generic verbs because they cannot access more specific content words [37,40]. These challenges in retrieval and speech planning lead to fragmented discourse, breaking speech into smaller, more manageable segments. This disruption results in a slower overall speech rate, evidenced by shorter verbal sequences in both duration and syllable number [40,41,22].

Research on the use of nouns, verbs, and pronouns in anomia has produced mixed findings. Anomia has been shown to impair word retrieval and production, impacting the frequency of use of nouns and verbs [42,43]. Individuals with anomia consistently generate more verbs than nouns [44]. Conversely, another study showed that fewer nouns were used with individuals relying more heavily on pronouns, while their verb usage remains similar to that of healthy controls [45]. Based on these findings, we implemented anomia as a speech biomarker for ADRD, which we detail in Section 3.

### 2.1.2 Acoustic-based features

In addition to content-based features, acoustic characteristics of speech such as prosody, pronunciation, and dysarthric speech provide critical insights into the motor and phonetic aspects of cognitive impairments. The following explores prior research on acoustic biomarkers and their potential role in assessing speech in dementia.

Broadly speaking, acoustic features are characteristics of sound that can be measured quantitatively. Key feature types for speech are pitch, loudness, duration, timbre, formants, harmonics, spectral features, and Mel-Frequency Cepstral Coefficients (MFCCs) [46]. Pitch refers to the perceived frequency of a sound, determining how high or low it sounds. Loudness is the perceived intensity or volume of a sound, while duration measures the length of time a sound lasts. Timbre distinguishes different sounds with the same pitch and loudness. Formants are resonant frequencies of the vocal tract that shape the sound of speech. The first two formats, F1 and F2, are particularly important in distinguishing vowel sounds [47]. Harmonics are frequencies that are integer multiples of the fundamental frequency, contributing to the richness of a sound. A common measure of harmonics is harmonics-to-noise ratio (HNR). Spectral features, such as spectral centroid and spectral bandwidth, describe the characteristics of the sound spectrum, which can capture the nuances or idiosyncrasies of an individual's manner of speech (e.g. rhythm). Finally, MFCCs are ubiquitously used in speech recognition research. They are meant to represent the short-term power spectrum of a sound and map it to the Mel scale, which can similarly capture nuances and idiosyncrasies of an individual's speech [48]. Of course, each key feature group has many forms of measurement therefore the number of possible acoustic features obtainable from speech is easily in the hundreds in number. As such, these features are used often in pathological speech analysis to aid in the treatment of various speech disorders.

Given such background research, we were interested here in generating **Prosody Change** and **Slurred Pronunciation** biomarker scores for dysarthric speech as part of a larger system for estimating the current cognitive health status of PLwD. Dysarthric speech occurs in dementia due to the loss of motor control of the vocal cords and muscles in the mouth, thought to relate to nerve cell failure [8, 11]. In the current body of literature, many types of acoustic features have been proposed to classify or detect dysarthric speech. In fact, many of these studies over the past decade extracted features based on some configuration of features first identified during the Interspeech 2009 emotion challenge [49]. Contained in many of those various configurations are acoustic features, which relate to pitch, amplitude, frequency, shimmer, etc. of speech, including acoustic spectral characteristics such as the aforementioned MFCCs. For instance, in previous research, the raw audio waveform was processed into MFCC vectors with a sliding window of 25ms and 10ms stride. The resulting sequence of MFCC vectors was then fed into a deep learning long-short-term memory (LSTM) network. Models using this approach performed well, achieving 0.90 AUC for a binary classification of dysarthria in individuals with cognitive impairment post-stroke [50]. Prosody and voice quality features also make up a substantial portion of the acoustic feature set in many research studies. For instance, Hernandez et al. 2009 used both types of features as input to SVM, Neural Network, and Random Forest models, achieving an overall accuracy of 75% in predicting dysarthria [51]. In addition to these types of acoustic features, there have been some studies exploring the use of glottal features, which are related to phonetic sounds specifically made in the back of the mouth or throat. For instance, Prabhakera & Alka 2020 explored this method in conjunction with the aforementioned acoustic parameters. Results of this study show glottal parameters may significantly improve classification of dysarthria compared to solely using acoustic features [52].

Outside of traditional extracted feature sets for ML methods, other research has utilized the raw speech data as input modeling data directly without first extracting separate features. For instance, one study proposed a method that used a deep neural network to learn feature extraction, normalization, and compression directly from raw speech data. The model was applied to sentence-level audio recordings to detect dysarthria and reportedly achieved 10% improvement accuracy over extracted feature-based models [53]. We describe our approach to Prosody and Pronunciation biomarkers based on human conversations with a robot in Section 3.

### 2.1.3 Turn-Taking as a Speech Biomarker

Conversational turn-taking is currently understood to involve both projection and reaction as turn "handovers" of active speaking occur during conversations. Listeners not only react to the end of the current speaker's utterance, but also project/predict when the current speaker will finish and prepare a response in advance [54,55]. Empirical evidence shows that people often begin responding within 200 ms of another speaker's turn ending, or even overlap the previous speaker slightly, even though fully formulating a response should theoretically take at least 600 ms [56]. Thus, for smooth turn-taking involving these near-instantaneous responses to be possible, the listener must not only anticipate what the speaker is going to say, but also when they are going to finish by using subtle clues embedded in the conversational discourse [57]. Those cues may be both verbal (e.g. slightly raised pitch near the end of a turn in English) or non-verbal (e.g. gestures, often culturally-specific).

Dementia has been shown to impair many of the cognitive domains integral to the turn-taking process, including memory, executive functioning, and semantic retrieval [58]. As those cognitive capabilities deteriorate, individuals with dementia may struggle to project the end of a speaker's utterance or formulate a timely response, leading to longer gaps during turn handovers, overlapping speech, and other conversational breakdowns [58]. In line with this, prior work demonstrates that ADRD can be reliably distinguished from healthy controls by quantifying temporal speech characteristics such as disrupted turn-taking instances, gap length, increased pause frequency, and shorter interpausal units [59,60]. Moreover, De Looze et al. (2021) reported moderate to high classification accuracies using automated temporal speech-based metrics of disrupted turn-taking, providing further evidence that these features can serve as practical speech biomarkers for detecting and monitoring dementia-related cognitive decline [61]. Building on these insights, we developed a **turn-taking biomarker** that quantifies overlapping speech events in real-time between a robot and human speaker, which we describe more in Section 3 below.

### 2.1.4 Multiple Speech Biomarkers

There is relatively limited research in the published scientific literature on integrating *multiple* biomarkers for dementia detection. Most existing studies have primarily focused on single biomarkers, relying on either text-based features [28] or acoustic features [62]. Moreover, those studies often focus on utilizing ML algorithms to predict dementia, rather than looking at the progression over time of symptoms like cognitive impairment. However, there are some studies that have attempted to combine acoustic, lexical, and disfluency features to train classification and regression models, demonstrating the potential of multi-modal approaches [63]. Others have employed cascaded methodologies to integrate data from multiple modalities, such as audio, text, and eye-tracking across various language tasks [63,64]. Additionally, some research has explored linguistic features extracted from clinical notes and structured non-speech data, highlighting the potential use of only text-based analysis (which obviously lacks acoustic features of spoken language) [65].

However, many of the aforementioned studies often rely on basic low-level linguistic and acoustic features (e.g. vocal shimmer) and do not extensively investigate the integration those features into more advanced biomarkers, such as slurred pronunciation, conversational coherence, alterations in overall grammar use, and disrupted turn-taking patterns. Furthermore, there is limited research on incorporating the more advanced biomarkers into a unified framework or integrating them into real-time conversational speech systems (e.g. onboard a robot) for continuous monitoring and early detection of cognitive decline. Addressing this gap presents a significant opportunity for future researchers to develop more comprehensive and nuanced models for detecting cognitive decline in ADRD. As such, we also evaluate a "comprehensive biomarker" score in this paper, which combines the individual biomarker scores into a single score.

## 2.2 Human-Robot Interaction (HRI)

As mentioned in Section 1, there has been much recent research on using social robots for dementia to assist both PLwD and their caregivers via human-robot interaction (HRI) [15]. Generally speaking in HRI, one of the most natural methods for communication between humans and robots is using conversation and/or SDSs, which make them natural avenues to explore speech biomarkers.

According to A. Russo et al. (2019), meaningful communication is a fundamental requirement that a social robot should possess when it encounters PLwD [66]. For this reason, researchers in HRI have specifically investigated the development of improving robots' communicative abilities through speech-based natural-language conversations from a number of angles. For instance, prior studies in HRI have included Socially Assistive Robots (SARs) as one of the main ways to communicate with PLwD to reduce loneliness and isolation of those individuals, which sometimes includes speech interaction (though not always of course)  SARs are a subcategory of robots that serve to provide social facilitation and companionship to users, with the goal of assisting through social interactions with users [67]. However, Zhou et al (2021) have argued in their studies that such robots should not be used to replace human companionship but rather to "augment" it, because PLwD prefer human assistance rather than robotic assistance. In that vein, other research has shown that using robots to enhance existing human companionship is more effective than replacing it in many cases [68].

Though most researchers have not tailored their studies based on  real-time responses in studies on HRI and dementia, Otaka et al. (2024) highlighted the potential effect of the  such "emotionally intelligent" real-time responses to user speech (e.g. affective signals such as happiness or stress/agitation) ) because it helps caregivers confirm the effectiveness of the robot and make better decisions about whether or not the PLwD should continue using the robot [69]. While communication is one of the main capabilities that a social robot should have, there are other criteria as well of course that going beyond what is conveyed (i.e. content of speech), such as trust and acceptability related to how things are communicated. Acceptability is defined as the robot willingly being a part of the user's life for long-term usage and assessing how well technologies are accepted by users as meeting their needs [70]. Similarly, trust can be defined as whether the users think the technology is reliable and paying attention to what they want and will respond in ways that make sense [71]. Using trust and acceptability in HRI is important because it help meet users' needs and increase user adoption of the technology.

Prior research on trust and acceptability shows that most researchers agree with the idea that robot's appearances should be based on user-centered design and (when speech is incorporated) should deliver human-like conversational styles [67,68,70]. Likewise, Koebel et al. (2022) found that technology derived only on expert insights (even if it converges with the other researchers' opinions) rather than user feedback during the design process can lead to challenges of user adoption, particularly if there are weaknesses regarding usability, desirability, and adherence of a given technology [72]. According to studies from A Russo et al. [67] and Whelan et al. [70], this was the case for a robot called iCat, where users preferred robots that were more socially expressive and that they could extensively communicate with in a free-form manner (which the iCat lacked at that time unfortunately). On the other hand, other researchers have claimed that dynamic social features - which include arm and hand gestures, body and head movements, vocal intonation and facial expressions – still remain of unclear value in HRI for providing effective and engaging assistance to users [73]. Given that robotic systems are constantly being improved to meet users' preferences in HRI, these aspects are important factors to consider when designing robots to detect speech biomarkers of dementia.

### 3. Methods

For this study we developed a "Dementia Speech System" designed to be deployed on conversational robots, which could calculate multiple speech biomarkers in real-time based on those conversations. An example of the future envisioned human-robot conversational scenario can be seen in Figure 1. We note however that the robotic platform being used in our research is the Buddy Robot (https://www.bluefrogrobotics.com/buddy-en), both a physical version and a 3D "digital robot" virtual avatar version that we created to resemble Buddy (see Figure 2, also video GIF in supplementary material). We also developed an accompanying mobile app "user interface" for users (PLwD and their caregivers) so that they could see the resulting biomarker data in real-time, as well as for displaying the robot virtual avatar if desired (see Section 3.2). We are currently conducting user testing of the mobile app user interface with human participants *in vivo* in lab settings. **However, this paper focuses primarily on the development of the Dementia Speech System and speech biomarkers**, rather than the accompanying mobile app or robotic platform.

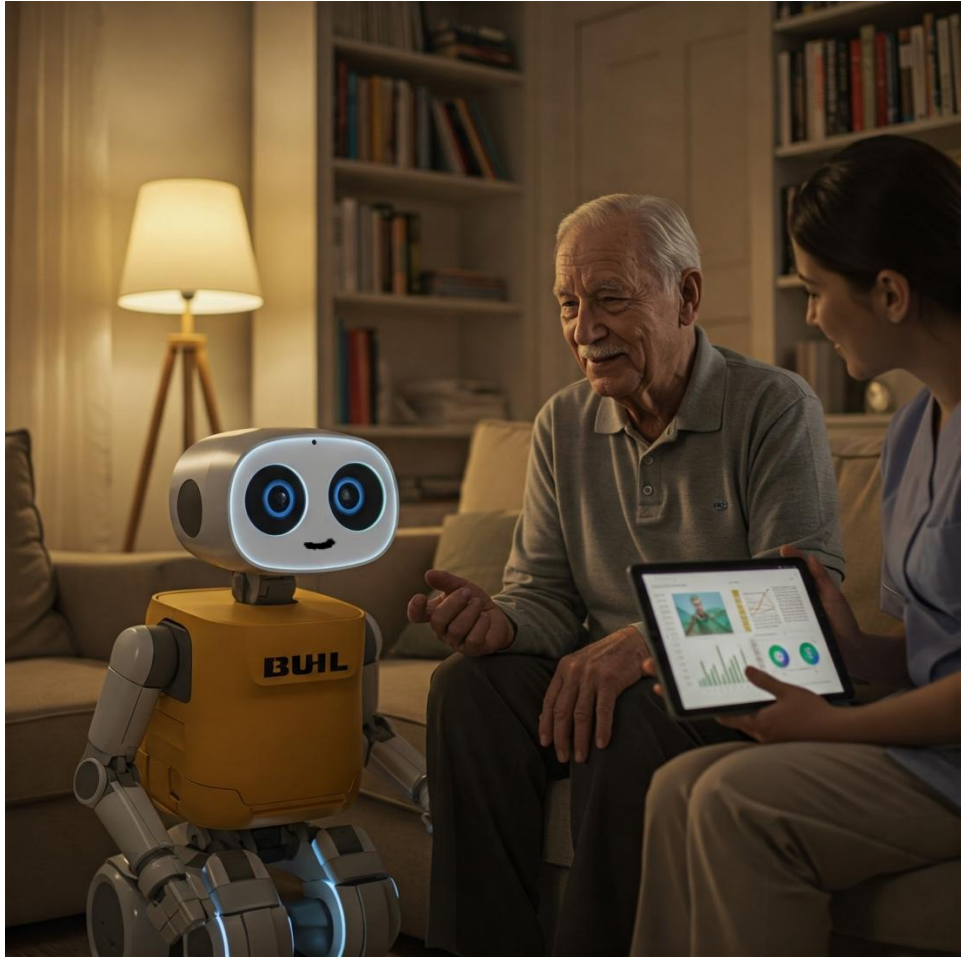

**Figure 1:** Envisioned future human-robot conversational scenario with older adult PLwD in their own homes along with caregivers using user interface on tablet, as generated by Google Gemini (https://gemini.google.com/app)

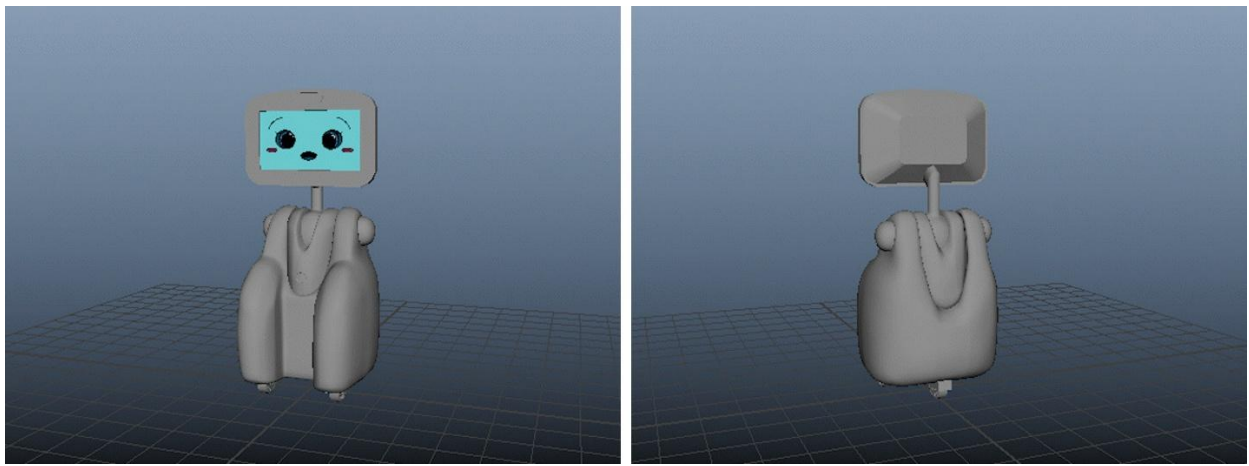

**Figure 2:** "Digital Robot" 3D virtual avatar that we created specifically to resemble the physical Buddy robot for this research (created in Autodesk Maya)

In this section, we first describe the technical components of the Dementia Speech System (backend system and frontend interface), the speech biomarkers deployed as part of the system, and finally the speech corpus data and analysis methods used for testing. In particular, we address some of the computational challenges of creating the system with low-latency that could be used during real-world conversations, where responses need to be generated within 1-1.5 seconds.

## 3.1 Backend Development

The backend of the Dementia Speech System as shown in Figure 3. It was designed as a modular and highly interactive architecture, enabling real-time speech responses for conversation and the extraction of six linguistic biomarkers to assess cognitive ability in people with dementia. We briefly describe some of the modules incorporated into the backend below, which at a high-level can be split into two parts: Python WebSocket Server and the Central Module (labeled "main function" in Figure 1). We describe the user interface later in Section 3.2.

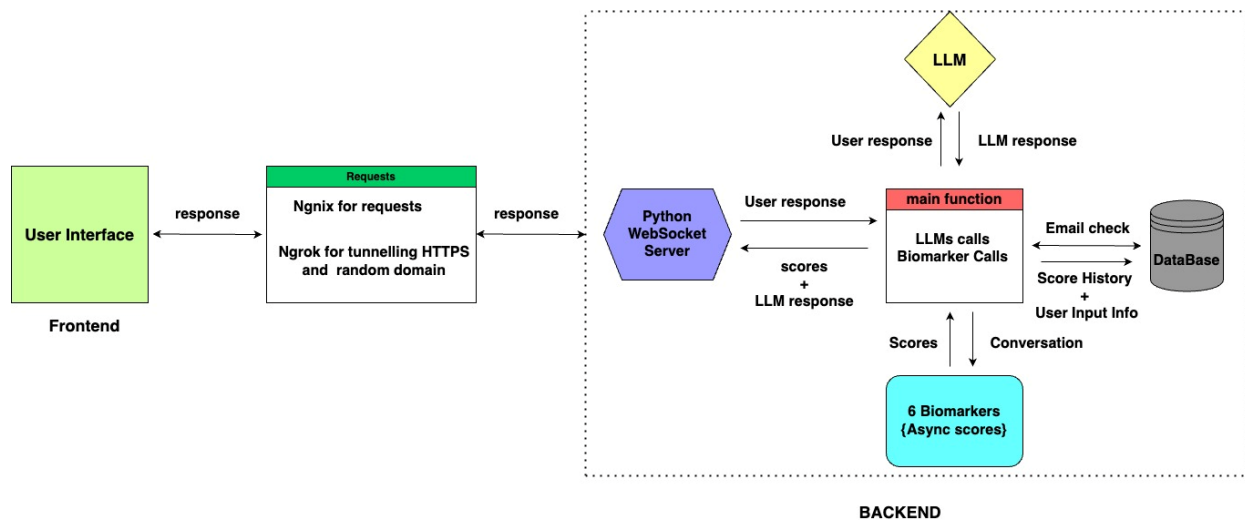

**Figure 3:** Overview of the "Dementia Speech System" backend (within dotted line box)

### 3.1.1 Python WebSocket Server:

The WebSocket server acts as the core communication layer between the frontend user interface and the backend processing modules. It receives user input (text or audio) as "requests" from the user interface, processes that incoming data to be handed off to the Central Module, and then returns structured responses to the user interface comprising both speech biomarker scores and speech utterances for the robot generated by the LLM component within the Central Module. Although the WebSocket Server does not contain as many sub-modules as the Central Module, it similarly handles multiple roles within the overall system and serves as a key link between the frontend and the rest of the backend that maintains the *computational efficiency* of the overall system. There are two key components of the WebSocket Server, described as follows.

- **NGINX and Ngrok Integration**: The backend leverages NGINX for handling requests and managing resource allocation efficiently. While this occurs mainly when requests are being picked up the WebSocket layer, it does affect downstream processing in the Central Module. The request from the frontend user interface first reaches NGINX, which acts as a reverse proxy and load balancer. Ngrok is utilized for secure HTTPS tunneling and creating dynamic domains for testing and deployment purposes that ensures secure communication over the internet for obvious data privacy reasons.
- **Interaction Workflow**: The speech system interaction begins at the frontend user interface, where the user (e.g. PLwD) interacts with the system through speech conversation. The spoken input is first converted into bytes to extract acoustic features and also transcribed into text using Azure's automatic speech recognition (ASR) **tool to derive linguistic features**. This processed input is then sent as a 'request' to the backend Python WebSocket server for further analysis. Before reaching the WebSocket server, the request is routed through NGINX, which manages incoming traffic and ensures proper redirection. The WebSocket server then waits for a signal from the Central Module, indicating it is ready to receive new data before passing it along for further processing.

### 3.1.2 Central Module

This Central Module (centered around the "main function" in Figure 1) orchestrates various sub-processes that lie at the core of the speech system, allowing speech utterances of the user to be detected, parsed, and biomarkers calculated, while at the same time facilitating responses from the robot. We describe those sub-processes below. We note also that the Central Module is not connected directly to the frontend user interface but rather communicates to the Python WebSocket Server described in Section 3.1.1.

- **Custom LLM Integration**: The user input is processed using a fine-tuned LLM, which provides contextual responses *specifically* trained on conversations with PLwD. This fine-tuned LLM serves as the core response engine for the speech system. To create that engine, we used as the base model the "Phi-3-Mini-4K-Instruct", which is a 3.8B parameter, lightweight, state-of-the-art open model trained on the Phi-3 dataset. It is part of the more general Phi-3 family of LLM base models, with the "Mini" lightweight version available in two variants (4K and 128K), representing the context length (in tokens) that it can support [74]. We fine-tuned the base Phi-3 model using Low-Rank Adaptation (LoRA) and 4-bit quantization to optimize performance while reducing memory usage. The fine-tuning was done using transcriptions from the Indiana dataset (described in Section 3.3 and 3.4 below), which were fed into the base model using the chat prompt format described below. Snippets of the recorded transcription in varying order were used as the human user input. This approach helped the base model understand the data more effectively after fine-tuning, as the base model was originally trained on this same data format (using general speech data, not dementia-specific).

### Chat Prompt Format

*</|System|>*
*Scenario Prompt. </|end|>*
*</|Human user|>*
*Question/Statement </|end|>*
*</|AI assistant|>*
*Generated Response </|end|>*

### Generic Example Prompt

*</|System|>*
*You are a socially assistive robot designed to respond to people with dementia in a friendly manner. </|end|>*
*</|Human user|>*
*Hello. How are you? </|end|>*
*</|AI assistant|>*
*Hello, I am fine, I am AI assistant, how can I help you? </|end|>*

However, for actual training we of course replaced the generic "human user" statements in the above example with snippets from the actual recorded conversations with PLwD.

The base model, Phi-3-Mini-4K-Instruct, was further modified with a custom tokenizer and quantization settings. The quantization technique included several specific settings for computational reasons: use of nf4 type, torch.float16 compute data type, and disabled double quantization. Training was conducted using parameter-efficient fine-tuning (PEFT) with LoRA to produce the fine-tuning layers, while freezing most of the original parameters of the base model. This approach ensures that the final custom model remains computationally viable for real-world applications, enabling it to deliver quick responses.

- **Database Integration**: The system includes a database to store user interaction history, past conversational topics with the user, computed biomarker scores, and log records of any errors in the speech system. We note that user interaction history is turned off by default for user privacy purposes but can be turned on within the scope of individual research studies if desired. This setup ensures security for the web application, session persistence, and support for longitudinal studies on cognitive decline.

- **Conversational Response Generation**: The transcribed input text is forwarded to our custom LLM model, which then uses the input to generate a meaningful and contextually relevant conversational response. The LLM output is also utilized to measure "semantic coherence" of user speech for calculating the Pragmatic Impairments biomarker (see Section 3.3). The generated utterance for the robot is sent back to the frontend user interface through the WebSocket layer, where it is converted to audio speech using Azure's text-to-speech (TTS). The entire "round trip" process (starting with user utterance to robot verbal response) has been purposely ***designed to occur in less than 1.5 seconds*** so as to maintain user engagement and facilitate natural interaction during the conversation with the robot.

- **Speech Biomarker Analysis**: While the robot conversational response is being generated, the user speech data is also simultaneously passed to six speech biomarker sub-modules, each designed to analyze specific features of the input (described in Section 3.3 below). These sub-modules assess critical indicators of cognitive impairment in ADRD, such as Anomia, Altered Grammar, Disrupted Turn-Taking, Pragmatic Impairments, Slurred Pronunciation, and Prosody Changes. The biomarker scores are calculated asynchronously to minimize conversational response latency and are updated every 5 seconds, given that we need different amounts of data to calculate different biomarkers. After the biomarker calculations, the six biomarker scores are sent back to the frontend user interface, where they are visualized graphically on a dashboard. This visualization allows for a comprehensive real-time understanding and analysis of the severity of cognitive impairment in PLwD for users, including caregivers of those with dementia and potentially clinicians as well. Additionally, the computed biomarker scores and their timestamps are stored in a database for longitudinal analysis and future research.

## 3.2 Frontend User Interface

The frontend user interface of our Dementia Speech System was designed as a Progressive Web App (PWA), which is a type of application that combines many desired features of web-based and native mobile apps into one package that can be deployed on both IOS and Android phones (or even desktop PC) [75]. PWAs are built using standard web technologies such as HTML, CSS, and JavaScript, but provide a common user experience across a variety of platforms (e.g. IOS, Android) that still feels similar to a native app within each platform. Since these applications are web-based, they eliminate the need for separate codebases on different platforms and ensure secure data transmission over HTTPS. This is well suited for our purpose of creating an interface and caregiver dashboard for our speech system as we eliminate the need for multiple codebases while ensuring a responsive and secure application. The application can also be easily installed on both a desktop or smartphone, which offers additional flexibility.

Our PWA is run using a 1) Django framework, 2) PostgreSQL database to store application data, and 3) standard coding tools (HTML, CSS, JavaScript) for displaying the visual frontend user interface. Additionally, Django channels are employed to handle the WebSocket Server connections for real-time communication between the frontend and backend. Python was chosen for the framework architecture because we had existing web socket code already written in Python for a related project, so it was easier to integrate that code into a new PWA without starting from scratch. Python is also very commonly used in backend programming so there is significant support and resources for implementing applications with it. Django was chosen as the application framework because of its versatility. It comes with a plethora of built-in features allowing us to develop quickly and adapt it to any change in our system architecture needed in the future. PostgreSQL was chosen for the database because of its strong security features and open-source platform, meaning the Dementia Speech System for conversational robots can be made widely available to the scientific community for research.

To convert the initial draft version of the user interface to a PWA, a "service worker" was added, which is a script that runs in the background, separate from a web page, enabling features like offline functionality, background sync (for data), and mobile push notifications [75]. Additionally, a "Web App Manifest" was included in the PWA, which is a JSON file providing metadata such as a name, icons, and starting URL. Such metadata allows the app to be installed on a home screen/desktop of a device, enabling easier access for users.

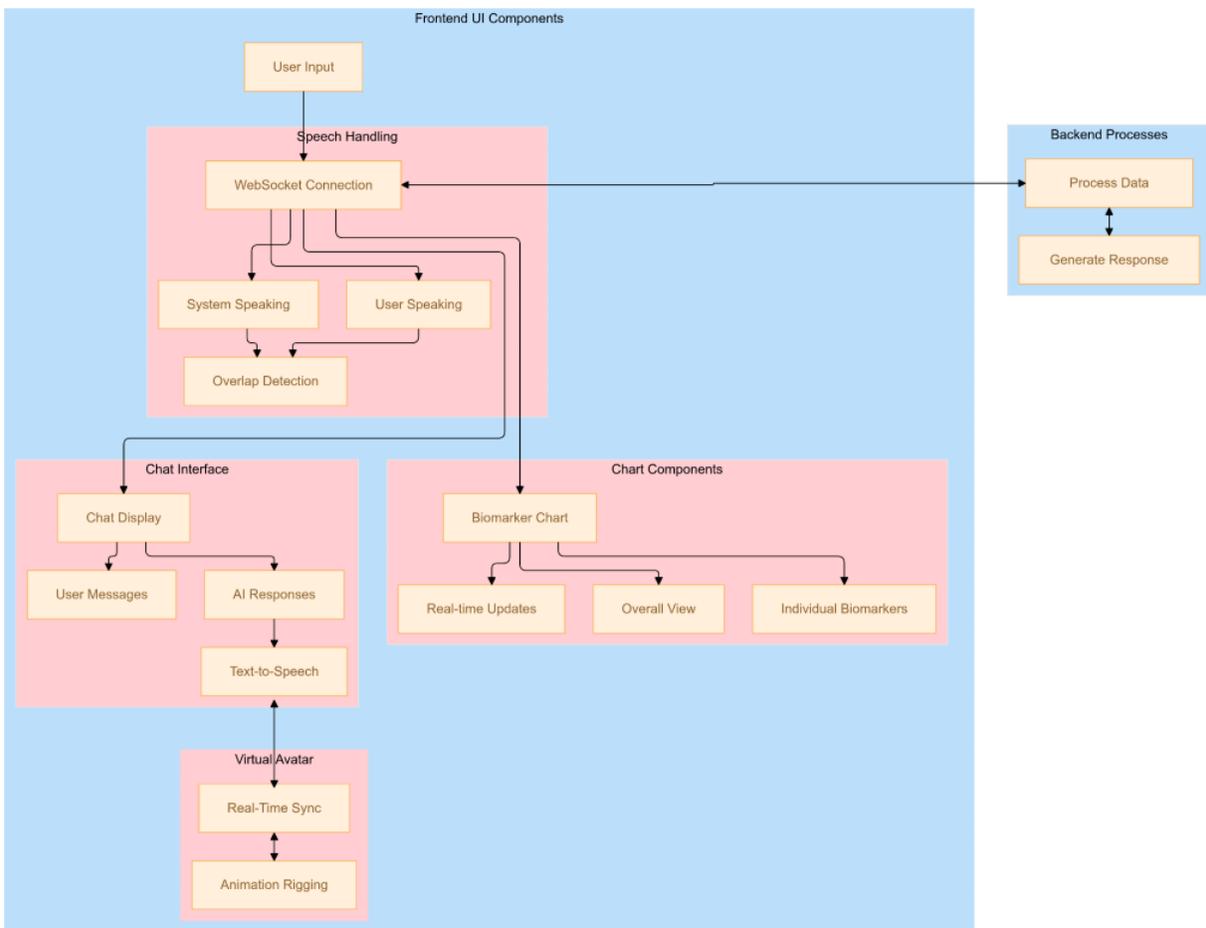

**Figure 4:** Overview of the "Dementia Speech System" frontend

Figure 4 illustrates the architecture of the frontend application components. User input, in the form of speech, is captured by the frontend, which buffers 0.25 seconds of speech and transmits it to the backend in JSON format. Concurrently, audio transcription via Azure's speech-to-text (STT) module is also generated and sent to the backend. The backend processes the data, generates a response from the speech model, and computes biomarker scores. This processed data is then transmitted back to the frontend mobile app, where the biomarker data is visualized graphically as "charts" in real-time. The frontend also displays both the user and system speech in the mobile app in a "chat" window on-screen for accessibility purposes (i.e. for those who are hard-of hearing). The mobile app also includes text-to-speech (TTS) functionality for the system speech, for situations where we may prefer to use an on-screen robot "virtual avatar" rather than a physical robot. It may not always be practical to deploy a physical robot for cost or logistical reasons, so having the option of using the virtual avatar to chat with on a mobile device holds benefits for both clinical use and research use.

### 3.3 Speech Biomarker Development:

Six speech biomarker scores were developed for real-time deployment in our robotic speech system based on two extant speech corpuses. Those corpuses were audio-only recordings from the DementiaBank dataset [76] comprising 306 samples from individuals with dementia and 242 samples from controls (without dementia), as well as the Indiana dataset comprising 27 audiovisual recordings of individuals with Dementia speaking with the QT Robot during Ikigai sessions (I.R.I.S.) [77,78]. To be clear, the Indiana dataset recordings were collected *prior* to the development of the Dementia Speech System here, so those studies did not involve the use of anything described above in Sections 3.1 or 3.2. Details of the HRI conversational setup in the Indiana studies can be found in [77,78], with an example in Figure 5. Using the Azure STT module and speaker diarization module (the latter to distinguish between different speakers in the same recording), we transcribed both datasets into text transcriptions to be used as the data. After initial development, we subsequently used those same 2 corpus datasets for testing by including clinical outcome scores not used during the development phase, as described in Section 3.4.

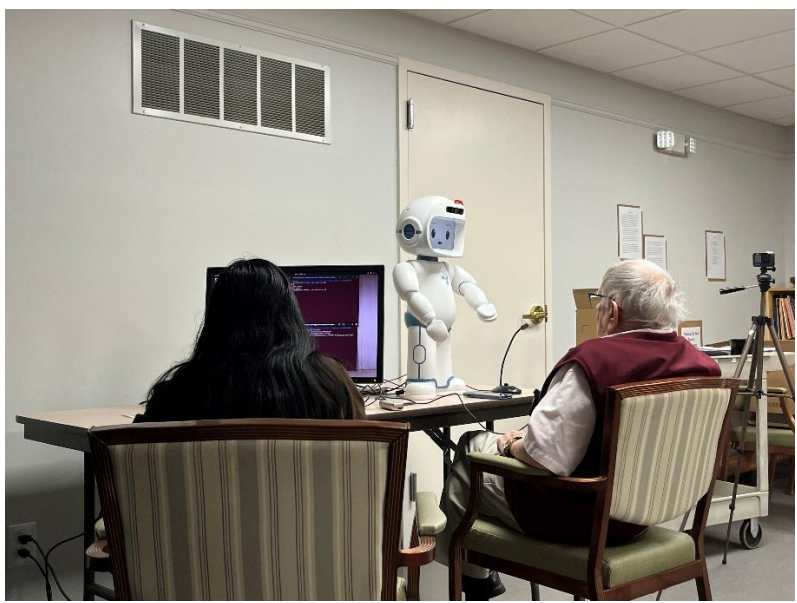

**Figure 5:** Example of the Indiana Dataset conversational scenario (QT Robot)

### 3.3.1 Altered Grammar:

Altered Grammar score was initially calculated by extracting various syntactic features and lexical features from the audio datasets using extracted transcriptions. These features provide insight into an individual's syntactic abilities when speaking, reflecting the level of complexity and structure of their language usage. Among the various possible syntactic and lexical features that could be calculated from the existing literature (see Section 2.1), 10 features were identified for the model that each appear to contribute significant amount towards determining the quality of an individual's spoken language, based on their weighted coefficients in a logistic regression model predicting dementia versus control (no dementia) in the DementiaBank dataset. Those features are listed below.

**Syntactic Features:** We extracted various syntactic features from the transcriptions using a syntactic tree structure of sentences and identified parts of speech (POS) tags produced by Stanford Parser, which is a natural language processing (NLP) parser. The resulting information can be used to capture the grammatical structure of the sentences. The proposed syntactic features derived from are as follows:

- **Coordinated Sentences:** Identified by the presence of coordinating conjunctions (CC) in the parse tree, these are sentences combining independent clauses with conjunctions like and, but, or, and yet.
- **Subordinated Sentences:** Measured by the frequency of subordinating conjunctions (S) such as because, although, while, and if, indicating dependent clauses.
- **Reduced Sentences:** Counted by detecting gerunds (VBG) and participles (VBN), these are subordinated sentences without explicit conjunctions.
- **Number of Predicates:** Extracted using a rule-based algorithm based on nominal verb dependencies, representing actions or states within sentences.
- **Production Rules:** Calculated by the number of unique context-free grammar production rules (e.g., NP → Det + N) derived from each individual's narratives.

**Lexical features:** The lexical features extracted in this study are derived from an approach presented in [28] which provides quantitative lexical complexity in each spoken narrative of PLwD.

- **Function words:** Includes prepositions, conjunctions, auxiliary verbs, and pronouns, identified via POS tags in the parse tree to assess grammatical coherence.
- **Immediate Word Repetition:** This feature captures the number of immediate word repetitions, referring to the consecutive repetition of the same word in the narrative.
- **Word Count:** The total number of words in the speech is computed, including both unique words and repeated instances.
- **Character Length:** The narrative's overall character length is measured to provide text size.

- **Unique Words:** The total number of unique words is calculated by removing repeated words from the overall word count.

We provide an analysis of some of the different ways we modeled these features for the Altered Grammar biomarker and the resulting performance of those models in the Results section below (Section 4.3). We note here that the final biomarker was based off a logistic regression model, where the score was calculated by using the 10 feature coefficients to compute a probability score via a linear combination of the trained coefficients ($\beta$ values) as represented in Equation 1 below.

$$f(x) = \beta_1 x_1 + \beta_2 x_2 + \cdots \ldots + \beta_n x_n \qquad (1)$$

### 3.3.2 Pragmatic Impairments:

For this biomarker, we calculated whether the PLwD is able to stay on-topic or not while having conversation, i.e. pragmatics. Pragmatic Impairment refers to the *inability* to stay on-topic. In order to capture that as a biomarker metric we used a local "semantic coherence" function [32], which computes how closely the human's utterances relates to the recent conversational topic during the human-robot conversation (or any conversation, for that matter). Essentially that function measures how much the current utterance relates to previous utterances using semantic similarity across the different words contained in those utterances. For instance, in a conversation about "seasons," if a PLwD says "I love the colorful leaves in autumn," the words "autumn" and "colorful leaves" are semantically related to "seasons", demonstrating strong topic coherence. Conversely, if the same topic prompts a response like "I enjoy eating pizza in winter", then although "winter" is related to "seasons", the shift mid-sentence to "pizza" may indicate a lower semantic coherence depending on the flow of the conversation and words used in the preceding utterances. The original code for calculating semantic coherence was developed in the R programming language by other researchers [32], so for our purposes here we converted the calculation into Python code to produce the biomarker score for pragmatic impairments.

### 3.3.3 Anomia:

Anomia significantly disrupts the natural flow of speech by making it difficult for individuals to retrieve specific words on-the-fly during conversation. This challenge often forces them to compensate by inserting filler words like "um" and "ahh," increasing the frequency of pauses, or fragmenting their speech into smaller segments. These disruptions can also alter their overall speech rate and change the proportion of nouns, pronouns, and verbs used, as shown in previous literature (see Section 2.1).

To calculate the anomia score, we first started by identifying and calculating filler words within speech samples. That was done by implementing a regex-based approach to identify filler words within the transcriptions. Our focus was on quantifying the occurrence of filler words, such as "um" and "ahh," as markers of speech hesitations and retrieval difficulties. These filler words were measured using the metric Filler words Per Minute (FPM), which provides a frequency-based representation of their occurrence within the speech. However, after some initial testing, we found that filler words alone were not sensitive enough to detect changes in real-time, and often result in a anomia score of 0 if used alone. Thus, we wrote a Python script using the SpaCy NLP library to extract nouns, verbs, and pronouns from the transcriptions. We then calculated their occurrences per minute to create metrics for Nouns Per Minute (NPM), Verbs Per Minute (VPM), Pronouns Per Minute (PPM), and overall words per minute (WPM) spoken by the PLwD. We then combined all those features (FPM, NPM, VPM, PPM, WPM) as represented in the equation below to compute a comprehensive anomia score.

$$Anomia\ Score = FPM * W_1 + NPM * W_2 + VPM * W_3 + PPM * W_4 + WPM * W_5 \quad (2)$$

Where W = weight. Note that for our anomia score, all feature weights were set to the same value of 0.2 (i.e. equal weight), but it is possible giving some features different weights could improve performance. We however did not test that idea here.

### 3.3.4 Disrupted Turn taking:

To compute the turn-taking biomarker, we analyzed the recorded speech transcripts for instances where one speaker interrupts the other, defined as overlapping speech (i.e. inter-pausal unit, or IPU) rather than waiting for the other person to finish speaking [54]. Such interruptions can be viewed as failures of proper turn-taking during a social interaction [54]. Moreover, that approach would allow us to calculate the biomarker in real-time during future deployment scenarios, averse to methods that require manual human processing to determine interruption occurrence. For our automatic process, we adapted a model we had originally designed for multi-modal turn-taking during human robot conversations in another study of ours [79], since the speech corpuses used here did not contain many instances of turn-taking between speakers.

For our purposes here, we began by analyzing the timestamps as an indicator of turn-taking instances, as those timestamps indicated the beginning/end of different speakers' utterances automatically based on the speaker diarization process. We then conducted an interruption analysis, identifying points where one speaker interrupts the other (who is still talking). This was done by identifying places where timestamps of utterances overlapped without any pause between (i.e. 0 milliseconds or less). To enable comparisons across sessions with varying speaking times and utterance counts, we normalize the interruption count by time (e.g., per minute). This interruption rate serves as the turn-taking biomarker, reflecting the extent to which speakers struggle (or succeed) in coordinating conversational turns.

### 3.3.5 Slurred Pronunciation and Prosody Changes

The openSMILE toolkit ComParE 2016 feature set was used as reference for extraction of the acoustic features to be used for training the prosody and pronunciation models [80]. The feature set is comprised of 65 features which are categorized into spectral, energy, voice-related, and cepstral. We used 59 of those features for pronunciation and 6 of them for prosody, as well as an additional 4 voice quality features that are shared between the pronunciation and prosody feature sets (i.e. unique to neither) [79,80]. To extract these features, a Python script iterated through each audio file and resampled it to a standardized sampling rate, followed by extraction of acoustic biomarkers via the openSMILE toolkit. The extracted features were segmented into 5-second non-overlapping chunks using a defined window size and hop length, ensuring temporal consistency across samples. Binary classification labels were generated for the DementiaBank transcriptions – 1 for dementia, 0 for control (no dementia) – and then those labels were aligned with the corresponding sample chunks. The resulting features from each video are concatenated together then reshaped into a 3-dimensional space (x, y, z), where x is the samples, y is the total chunks from each sample, and z is the features. This maintains consistent feature dimensionality and temporal resolution across all processed audio samples, even in cases where total recording length differs across samples.

Initially, two random forest models were trained using the prosody and pronunciation feature sets. Those random forest models could output a probability from 0 to 1, which determines the predicted classification (dementia or no dementia) but could also be used as a continuous valued biomarker indicating the "probability" of dementia in that sample. Given the known association of slurred pronunciation and prosody changes with cognitive impairment in dementia (see Section 2), we can then take that probability as an indicator of cognitive impairment to be used as the biomarker score.

We note that we did experiment with some other modeling methods for prosody and pronunciation biomarkers (e.g. logistic regression) as well as experiment with using feature selection to create smaller subsets of input features and reduce noise in the input dataset. We discuss the results of such experimentation in Section 4 below.

### 3.4 Testing & Analysis of Speech Biomarkers on Existing Speech Corpuses

To test our Dementia Speech System, we evaluated the performance of the speech biomarkers that are incorporated into it on two speech corpuses: DementiaBank and Indiana dataset (see descriptions in Section 3.3). That included 5 of the biomarkers, leaving out the turn-taking biomarker as there were not enough conversational hand-offs (turns) in either dataset to calculate that reliably. Rather, both datasets primarily featured the PLwD speaking, with only occasional utterances from the other party.

For the DementiaBank dataset, it includes a clinical outcome score for dementia known as the mini mental state exam (MMSE) score that can be used for evaluation by calculating correlations of the MMSE with the individual biomarker scores as well as with an overall "composite score" that combines all 5 biomarkers into a single score by computing their average. Ideally those correlations would be *negative* values closer to -1 than 0, as for the MMSE lower values indicate higher cognitive impairment levels. That approach could also potentially allow us to establish thresholds for each biomarker about what constitutes a "good" vs. "bad" biomarker score, by distinguishing between mild impairment cases vs. moderate cases vs severe cases using the MMSE. We analyzed such potential thresholds as well in our results. We also note that though we did use portions of the DementiaBank dataset for training models for some of our biomarkers, that training set did *NOT* include the MMSE scores. Thus, the MMSE can be considered "unseen data" in this situation.

For the Indiana dataset, there were no MMSE scores available unfortunately as that was not the purpose of that study [77,78], so we could not calculate correlations like above. Rather, the Indiana dataset constituted audiovisual recordings of people with mild dementia **conversing with a robot**, so the aim of that analysis was to compare whether the produced biomarker values on the Indiana dataset were similar to the DementiaBank dataset, whereas the latter were recordings of people with dementia **conversing with a human clinician**. Ostensibly, the biomarker scores should be similar across both datasets for each biomarker (both the average values and range), unless there was some difference due to the severity of cases or the conversational scenario (e.g. talking with a robot vs. talking with a human). We investigate that question in our results.

The process for calculating the biomarker scores on both datasets is detailed below, which is slightly different depending on if we are calculating acoustic-based biomarkers (prosody, pronunciation) or content-based. We have Python code that can be modified for use with any dataset of audiovisual recordings with people with dementia, which can be shared with other researchers upon request.

## Steps for Generating Prosody & Pronunciation Biomarker Scores:

1. Set up the Python script (e.g., generate_scores.py,audio_processv1.py) in folder location with dataset videos. Change any hard-coded file paths
2. Ensure the Python environment is set up correctly (install relevant libraries, .pkl file models, etc.) in your terminal.
3. In the generate_scores.py script, modify the conditional statements to specify which biomarker to calculate (Prosody or Pronunciation):
    o If generating Prosody scores, make sure the Pronunciation portions of the code are turned off (e.g. commented out), and vice versa.
    o Adjust other parts of the Python code based on your local environment, if needed
4. Generate scores, results will save to a .txt file for later use:
    o To not overwrite previous results, make sure to change any output file names
5. Repeat steps 4b to 5 until all scores are generated.
6. At the end, you will have generated scores for both biomarkers, which can be used for further analysis

**Steps for Getting the Grammar, Anomia, and Pragmatics Scores:**

1. Setup the relevant Python scripts, similar to prosody/pronunciation above. Change any hard-coded file paths
2. Set up your Python environment in your terminal (install relevant libraries, etc.).
3. Ensure you have the scripts labeled audio_processv1.py, transcribe_audio.py, and main.py:
   - transcribe_audio.py is required to generate transcriptions for these biomarkers, but all three scripts are needed to complete the process.
4. Run transcribe_audio.py in the terminal to generate transcriptions in an Excel format automatically.
   - Make sure the transcription output is in the same environment as the code's location to run main.py and obtain feature scores. If not, update python code to your local path
5. Before running main.py, ensure that the conditional statement (at the bottom of main.py) is set to the dataset directory's location. Relative paths are useful here for re-use later
6. Run main.py to generate a features.xlsx file, which will include scores for Grammar, Anomia, and Pragmatics as a single file.
   - To not overwrite previous results, make sure to change any output file names

## 4. Results

As a preliminary evaluation of our robotic Dementia Speech System, we analyzed the performance of the 5 speech biomarkers that are incorporated into the system using two speech corpuses: DementiaBank and Indiana dataset. That did not include the turn-taking biomarker for reasons mentioned in Section 3.4. We describe the details of the analysis process in Section 3 above, and here describe the results. We note that for a few biomarkers, we tested multiple versions of the same biomarker for comparison. We include that analysis at the end of the results below as well.

### 4.1 DementiaBank Analysis

The overall correlations of the speech biomarker scores with MMSE scores in the DementiaBank dataset can be found in Table 1. As can be seen along the bottom row of that table, the correlation of composite biomarker score was roughly -0.55, which is in line with other state-of-the-art scores reported in the recent literature falling in the negative 0.5-0.6 range and could be considered moderate correlation [81]. The negative correlation indicates here that higher biomarker scores are associated with lower MMSE scores (i.e. greater cognitive impairment) as hypothesized, though the biomarkers could easily be inverted if desired to make lower values signify "worse". Regardless, of primary interest here is that **the composite score of the combined speech biomarkers was much higher than any individual biomarker**, which suggests that approaches utilizing multiple biomarkers that capture different aspects of speech changes due to cognitive impairment in dementia may be a promising path forward for conversational robots.

**Table 1:** Speech Biomarker Correlations with MMSE scores of PLwD Individuals (DementiaBank)

|  | Grammar | Pragmatics | Anomia | Pronunciation | Prosody | Composite |
|---|---|---|---|---|---|---|
| **Min** | 0.0000 | 0.3378 | 0.1330 | 0.3550 | 0.2723 |  |
| **Max** | 0.9218 | 0.8444 | 0.9230 | 0.5400 | 0.6461 |  |
| **Avg** | 0.4366 | 0.5776 | 0.4069 | 0.4454 | 0.4428 | 0.4594 |
| **StdDev (avg)** | 0.2055 | 0.0908 | 0.1214 | 0.0098 | 0.0535 | 0.0629 |
| **Correlation w/ MMSE** | -0.4760 | -0.3797 | -0.2420 | -0.1692 | 0.0320 | -0.5515 |

The one speech biomarker that did not really perform that well despite our efforts was Prosody. However, we note that although it did not perform well individually, if removed from the composite score then that composite score's correlation with MMSE dropped significantly, down around -0.5. That reinforces the notion that speech biomarkers for dementia may be a case where "the whole is greater than the sum of their parts" when it comes to multiple combined biomarkers.

We were also interested in calculating thresholds for what constitutes a "good" vs. "bad" score for each biomarker in terms of cognitive impairment level (none, mild, moderate, severe). To do so, however, would require that the values were significantly different across different levels, and by extension the average values and min/max values for each level would not have a high degree of overlap. We tested that by breaking the DementiaBank dataset into groups based on severity level as determined by established MMSE cutoffs [82]. We then ran independent samples t-tests ($\alpha=0.5$) to test for differences across different groups, with the results shown in Table 2. We also include the MMSE correlations and min/max biomarker values for each group as tables in the Appendix in the online supplementary material, for interested readers.

**Table 2:** T-tests of the Speech Biomarkers for detecting Cognitive Impairment Severity Level (*= statistically significant)

|  | Grammar | Pragmatics | Anomia | Pronunciation | Prosody | Composite |
|---|---|---|---|---|---|---|
| **Moderate-Severe** | 0.97063 | 0.04314 | 0.79838 | 0.09722 | 0.61628 | 0.35248 |
| **Mild-Severe** | 0.13786 | 0.01811* | 0.65270 | 0.01909* | 0.29265 | 0.07451 |
| **Mild-Moderate** | 0.02977* | 0.53792 | 0.64749 | 0.02735* | 0.32776 | 0.07485 |
| **None-Severe** | 0.00003* | 0.00000* | 0.03630* | 0.01573* | 0.50207 | 0.00000* |
| **Mod/Sev vs Mild/None** | 0.00000* | 0.00003* | 0.00270* | 0.01733* | 0.43190 | 0.00000* |

As we can see in Table 2, our speech biomarkers were able to detect significant differences when the groups were further apart in terms of severity (e.g. none vs severe, none/mild vs moderate/severe), particularly for the composite score and several individual biomarker scores. The t-test significance levels were less than 0.001 in the composite case. When the groups were closer together in terms of severity (e.g. mild vs. moderate, mild vs severe), the overall composite score was just beyond statistical significance at 0.07, though a couple of the individual biomarkers scores attained significance for some cases (e.g. altered grammar, pragmatic impairments, pronunciation). None of the biomarkers nor the composite score could distinguish moderate cases from severe cases. Our overall interpretation of these results was that **while using multiple speech biomarkers for dementia may not be sensitive enough to detect small differences in cognitive impairment, larger differences in impairment seem to be easily detectable**. It may also be possible to improve the sensitivity of multiple speech biomarkers for conversational robots through further research allowing them to capture subtle changes in cognitive impairment, which could be a boon for the use of such robots for clinical purposes.

## 4.2 Indiana Dataset Analysis

Following up the DementiaBank analysis of our speech biomarkers, we also analyzed the same biomarkers using the Indiana dataset. As that dataset contained no MMSE scores, we could not calculate correlations like above. Rather, we instead compared the scores from the Indiana dataset with the DementiaBank dataset to see if the scores were similar or not across both datasets, since the Indiana dataset involved PLwD conversing with a robot while DementiaBank involved PLwD conversing with a human clinician. The results can be seen in Table 3.

**Table 3:** Speech Biomarker Correlations of PLwD Individuals during Human-Robot Interaction (Indiana Dataset)

|  | Grammar | Pragmatics | Anomia | Pronunciation | Prosody | Composite |
|---|---|---|---|---|---|---|
| **Min** | 0.5172 | 0.4431 | 0.0749 | 0.3150 | 0.0000 |  |
| **Max** | 0.8496 | 0.7572 | 0.8620 | 0.5700 | 1.0000 |  |
| **Avg** | 0.7582 | 0.6186 | 0.3565 | 0.4716 | 0.4616 | 0.5507 |
| **StdDev (avg)** | 0.0540 | 0.0638 | 0.1369 | 0.0028 | 0.0708 | 0.0500 |
| **Correlation w/MMSE** | N/A | N/A | N/A | N/A | N/A | N/A |

Comparing Table 3 with Table 1, we can see that the average values of the biomarkers in the Indiana dataset were generally higher than those seen in the DementiaBank dataset across all speech biomarkers (except Anomia), as well as the composite biomarker. In particular, Altered Grammar and Pragmatic Impairments were notably higher. There were also differences in the "spread" of values as indicated by the min, max, and standard deviation of each biomarker in both datasets (except Anomia). We attribute that possibly to differences in the nature of the study populations. The Indiana Dataset contained only PLwD from a memory care facility with moderate levels of dementia during longer conversations [77], but the DementiaBank dataset included mainly short clips of a wide variety of cases from mild to severe. However, that is difficult to assess here precisely given that the Indiana dataset lacked MMSE scores.

Another potential interpretation of those differences is that speech interactions with a robot may be fundamentally different somehow than conversations with another human, creating different "conversational scenarios" and thus producing different biomarker scores. For instance, the nature of the conversational activity in the Indiana dataset involved interactive story-telling activities with robots around objects like photos. The recordings were relatively long as well, leading to more "ebb-and-flow" in the conversations. Meanwhile, the DementiaBank dataset was very different in nature, often comprising short clips of "monologues" about a specific topic from the PLwD briefly interspersed with comments from a human clinician, i.e. less interactive and less ebb-and-flow to the conversation. Since biomarkers are calculated over time, more ebb-and-flow in conversations may result in less complex grammar variation or need to coherently stay-on-topic (i.e. pragmatics), since the conversation can just "die off" then an entirely new conversation starts a few moments later after a brief pause (i.e. like mini-conversations throughout the day, rather than non-stop talking). It is plausible that things like prosody and pronunciation would not be affected by that as much.

Regardless, the takeaway from these results seems to be that **speech biomarker values may be affected by the conversational scenario**, and what occurs in more naturalistic environments at home with robots will be quite different than what occurs with clinicians in the clinic. What we might consider "good" in one scenario would not be the same as good in a different scenario. It is an intriguing idea, but more research is needed on this topic.

### 4.3 Individual Biomarker Development

A few of the biomarkers required more iterative model development during the research, in particular Altered Grammar, Slurred Pronunciation, and Prosody. We provide the results of that work in this section.

### 4.3.1 Altered Grammar Modeling

To model Altered Grammar, we evaluated various ML classification methods based on grammar-related linguistic features for distinguishing people with dementia from controls (no dementia) in the DementiaBank dataset during the development process, including Decision Trees, SVMs, Random Forests, Gradient Boosting, K-Nearest Neighbors and Multiple Logistic Regression (MLR) [83,41,28]. Among these, the MLR model was chosen for the final model as it not only achieved good performance metrics (Accuracy: 0.745, Precision: 0.775, Recall: 0.821, F1-Score: 0.797) relative to other methods Random Forest (Accuracy: 0.745, Precision: 0.791, Recall: 0.791, F1-Score: 0.791) as illustrated in Table 4, but also demonstrated the highest recall. This higher recall is crucial for detecting dementia symptoms (e.g. cognitive impairment), since it ensures that a greater proportion of actual dementia cases are correctly identified, minimizing the risk of false negatives. Additionally, MLR was selected for its ability to provide interpretable feature coefficients and good performance with smaller feature set, as shown in Table 5. Those coefficients are essential for calculating the Altered Grammar score, which predicts the probability of cognitive impairment based on the syntactic and lexical features. For the results in Table 4 and 5, the MLR model was trained and evaluated using 10-fold cross-validation on DementiaBank samples.

**Table 4:** Performance Metrics of Various ML Classification Methods in Differentiating Dementia Patients from Controls with the Dementia Bank Dataset

| Model | Accuracy | Precision | Recall | F1-Score |
|---|---|---|---|---|
| Multiple Logistic Regression | 0.745 | 0.775 | 0.821 | 0.797 |
| Support Vector Machine | 0.736 | 0.799 | 0.791 | 0.785 |
| K-Nearest Neighbors | 0.691 | 0.770 | 0.701 | 0.734 |
| Decision Tree | 0.655 | 0.723 | 0.701 | 0.712 |
| Random Forest | 0.745 | 0.791 | 0.791 | 0.791 |
| Gradient Boosting | 0.709 | 0.769 | 0.746 | 0.758 |

**Table 5:** Coefficients derived from a Multiple Logistic Regression model trained using 10 linguistic features of altered grammar

| | Features | Coefficients (Weights) |
|---|---|---|
| 1 | Co-ordinated Sentence | 0.469133 |
| 2 | Subordinated Sentence | 0.140325 |
| 3 | Reduced Sentence | -0.773910 |
| 4 | Predicates | 0.304633 |
| 5 | Production Rules | 0.023355 |
| 6 | Function Words | -0.115484 |
| 7 | Unique Words | 0.040963 |
| 8 | Total Words | 1.238682 |
| 9 | Character Length | -1.814850 |
| 10 | Immediate Word Repetitions | 0.707059 |

### 4.3.2 Pronunciation and Prosody Modeling

Pronunciation and Prosody biomarkers were developed alongside each other, as they both rely on acoustic features derived from the same processing in OpenSmile. We attempted several iterations of both biomarkers during that process. Those iterations included different models (e.g. random forests, logistic regression) to predict dementia versus control (no dementia) in the DementiaBank dataset, as well as altering some model parameters and using feature selection to reduce the number of features used. We were able to make some minor improvements to the individual Pronunciation biomarker score relative to its prediction accuracy by reducing the feature set for modeling to only 18 features out of the original 59 (see Section 3.3.5). It seemed that using all 59 caused issues due to noise in the data. However, we were never able to improve the individual Prosody biomarker score relative to its prediction accuracy. It is possible future research may be able to address that, but it remains unclear as of now.

We also considered dropping either Pronunciation or Prosody (or both) out of our composite score, but we discovered that omission of either reduced the correlation with MMSE. As such, we retained both in that composite score calculation.

## 5. Discussion
## 5.1 Summary of Results

This paper details the development of a conversational robot that can extract speech biomarkers of dementia in real-time during conversations with people with dementia, as well as how that information can be delivered back to caregivers of PLwD in a mobile app interface. The paper discusses the technical details of such development, including both the speech biomarkers and the customized LLM-based speech system that enables the conversation. As shown throughout the paper, this kind of conversational robot system needs to rely on many different components on both the backend and frontend to be computationally tractable in the real world.

The paper also provided an analysis of the resulting speech biomarkers on two different datasets that comprise audio-visual recordings of conversations of people with dementia. The first dataset (DementiaBank) included recordings of PLwD conversing with human clinicians, while the second dataset (Indiana) included recordings of PLwD conversing with a robot. That analysis showed our composite biomarker score from the conversational robot was similar to other state-of-the-art scores reported in the recent literature. We also found that our speech biomarkers from the conversational robot were able to significantly distinguish between PLwD with different levels of cognitive impairment (e.g. mild vs severe) when the differences in severity were larger, but was not sensitive enough to detect small differences. That suggests more work is needed.

We also found some notable differences in the speech biomarker scores between the two datasets, with the Indiana dataset generally having higher scores (indicating greater average cognitive impairment) and more variability in the "spread" of scores. That may suggest that the **conversational scenario** may affect speech biomarkers even if the underlying calculations for each biomarker are the same, e.g. scores detected during PLwD conversations with a robot at home would be fundamentally different than PLwD conversations with a human clinician at a clinic, though we found it may depend on which speech biomarker is used (e.g. altered grammar vs. slurred pronunciation). That has immense implications for the deployment of speech biomarkers in the real world, which we discuss more below.

We are currently conducting user testing of the system with human participants *in vivo* in lab settings. That includes having them interact with both the Dementia Speech System itself, as well as the mobile app user interface that shares real-time data about the speech biomarkers with PLwD and their caregivers. We plan to publish that in future work to shed more light on the findings here.

## 5.2 Research Implications

### 5.2.1 Implications for robots & LLMs for Dementia care

As mentioned in Section 5.1 above, one of the interesting things that came out of this research was that the conversational scenario may affect the speech biomarker scores in significant ways, which implies that what we might consider a "good" or "bad" score in one context is not necessarily universal to all contexts. In the research here for instance, we had two PLwD speech datasets from different conversational scenarios, where both the setting (at-home vs in-clinic) and conversational partner (robot vs clinician) varied. Given that speech biomarkers for dementia are envisioned as a tool for real-world deployment in the future [12], that finding suggests we may need to radically re-think how we conduct research on the topic of speech biomarkers. Data generated in a controlled lab study on speech biomarkers may not be directly applicable to real-world use cases, which makes sense given what we know from other research fields [84,85]. However, it is important that we are cognizant of that fact in speech biomarker research. That is true not only for dementia, but also for other use cases of speech biomarkers such as autism, mental health disorders, and more broadly biomarkers of cognitive impairment that occurs during general healthy aging.

We should also note here that, while this study focused on speech biomarkers for conversational robots, the findings in this paper apply to conversational agents that may be deployed on many different devices in the future, e.g. smartphones, wearables, internet-of-things (IoT) devices, and other on-screen virtual avatars [86]. There continues to be a growing interest in adding speech interactions to those kinds of technologies averse to clicking buttons or typing as in the past [87], and, moreover, people are becoming increasingly comfortable with virtual interaction with characters on a screen or other humans (e.g. Zoom) in the post-COVid era [88,89]. Suffice it to say, research on speech biomarkers has implications for conversational "virtual robots" that may inhabit a smartphone or wearable as a way to extract useful health information from those devices in real-time without users explicitly having to enter such information in a tedious manner.

### 5.2.2 Clinical Implications

There are a number of clinical implications to this research, both within the clinic and outside of it. First of all, there is the obvious potential of speech biomarker data from everyday interactions to be fed back to clinical data systems, e.g. electronic health records (EHRs), which may provide a more holistic view of a patient to augment data collected during clinical visits [90]. It is well-known that there is a *limit* to what can be done with EHR data alone, as it only captures information at specific points in time [91,92]. Thus, expanding the information available in-clinic in a way that requires minimal user effort is advantageous.

It is also plausible to imagine telehealth applications that involve a conversational robot physically on-site with a patient while the doctor or clinician is virtual. Such a scenario is not that different than some of the recordings in the Indiana dataset we used here in this study, where a clinician/researcher was present while the robot and PLwD conversed (i.e. triadic interaction) [77,78]. In that case, the speech biomarkers can still be elicited from the PLwD, though we may need to consider the impact of a triadic interaction scenario on the biomarker scores, as mentioned in Section 5.2.1 above. That is a question for future research.

Finally, there has been interest for many years in creating "virtual nurses" and other agents to help patients remember important details during intake/discharge, scheduling appointments, medication reminders, etc. [93,94]. Speech biomarker research has implications for that as well. For instance, a common situation in hospital settings is to receive a phone call from a patient about scheduling or discuss some issue with their care team (e.g. nurse). Those conversations have traditionally been thought of as "support services" that are adjunctive to healthcare, but if we apply the same speech biomarkers proposed here to those conversations, then it may be possible to extract pertinent health-related information from them. Such information could be used later to make clinical decisions for treatment, diagnosis, and so forth, not to mention potentially as data for clinical decision support (CDS) systems [95]. The ability to collect such data seamlessly using natural interaction (conversations) could be transformative for clinical care, enhancing both the clinician's efforts while also enhancing the patient experience.

## 5.3 Limitations

There are a number of limitations to this study. The first one being that while we had two relevant datasets to analyze the effectiveness of our speech biomarkers on, those two datasets don't necessarily capture every possible conversational scenario that may occur between a PLwD with a robot or human clinician. In particular, there are a multitude of scenarios that can occur in the real-world, and one ongoing research challenge for the community is to figure out how to test conversational robots in a wide array of scenarios. Moreover, more publicly available datasets of human-robot conversations from those with PLwD would be very helpful, particularly ones that include outcome scores like MMSE to benchmark speech biomarker performance against. That limitation is, of course, not one that this study was designed to address, but regardless one that as a scientific community we need to address going forward.

A second limitation is that, while we have done some small-scale user testing of the Dementia speech system developed here during a series of participatory design (PD) workshops with 10 participants, we have not yet done large-scale user testing in user homes. Large-scale testing is planned for later, and we are publishing the PD workshop results in a separate paper focused on that, whereas this paper focuses mainly on the technical development of the system. Nonetheless, it is a current limitation at this point, but one that we are actively addressing in ongoing/future research.

Finally, another limitation is in the shear number of possible speech biomarkers and/or speech characteristic features that we could utilize. We settled on 6 biomarkers here for our system based on the existing scientific literature, but there is an endless array of possible combinations and it is not immediately clear when using multiple biomarkers what the optimal combination may be. Moreover, when using multiple speech biomarkers in real-time, we have to address the issue of computational tractability, since a system calculating biomarkers that causes a lag or delay in the conversational response from the robot typically creates awkward interactions with humans. We spent quite a bit of time dealing with the computational tractability during our speech system development, managing to get the "round trip" conversational responses to reliably occur in under 1.5 seconds. However, this is a limitation of current technology that needs to be addressed for real-world healthcare applications, both in terms of calculating speech biomarkers but also in terms of the underlying LLM and deep learning models that are used to generate them [96,97]. Power or accuracy is not the only thing that matters for such real world applications, but also computational *efficiency*.

### Acknowledgements

This research was funded by a University Research Council (URC) grant from DePaul University, grant award # 9474 (project 602448, to C.C. Bennett). We also received funding for parts of this work from Toyota Research Institute. We would also like to thank our collaborators at Jill's House, Dementia Action Alliance (Jan Bays), and Rush University Medical Center in Chicago.

### Author Contributions

Writing – original draft: CCB, RP, YHB, DI, AM, EW. Writing – review and editing: All authors. Conceptualization: CCB, SS. Funding acquisition: CCB, SS. Methodology: All Authors. Investigation: RP, YHB, DI, AM, EW, LJH. Software: RP, YHB, DI, AM, CCB. Formal analysis: RP, YHB, DI, AM, EW. Supervision: CCB.

### References

1. Alzheimer's Association (2023). 2023 Alzheimer's Disease Facts and Figures. *Alzheimer's & Dementia*. 19(4).
2. Matthews, K. A., Xu, W., Gaglioti, A. H., Holt, J. B., Croft, J. B., Mack, D., & McGuire, L. C. (2019). Racial and ethnic estimates of Alzheimer's disease and related dementias in the United States (2015–2060) in adults aged$\geq$ 65 years. *Alzheimer's & Dementia*, 15(1), 17-24. https://doi.org/10.1016/j.jalz.2018.06.3063
3. Stara, V., Vera, B., Bolliger, D., Rossi, L., Felici, E., Di Rosa, M., ... & Paolini, S. (2021). Usability and acceptance of the embodied conversational agent Anne by people with dementia and their caregivers: exploratory study in home environment settings. *JMIR mHealth and uHealth*, 9(6), e25891.

4.  Oewel, B., Ammari, T., & Brewer, R. N. (2023). Voice assistant use in long-term care. In *Proceedings of the 5th International Conference on Conversational User Interfaces (CUI)*, (pp. 1-10).

5.  Ruggiano, N., Brown, E. L., Roberts, L., Framil Suarez, C. V., Luo, Y., Hao, Z., & Hristidis, V. (2021). Chatbots to support people with dementia and their caregivers: systematic review of functions and quality. *Journal of Medical Internet Research*, 23(6), e25006.

6.  Pacheco-Lorenzo, M. R., Valladares-Rodríguez, S. M., Anido-Rifón, L. E., & Fernández-Iglesias, M. J. (2021). Smart conversational agents for the detection of neuropsychiatric disorders: A systematic review. *Journal of Biomedical Informatics*, 113, 103632.

7.  Ruitenburg, Y., Lee, M., IJsselsteijn, W., & Markopoulos, P. (2024). Seeking Truth, Comfort, and Connection: How Conversational User Interfaces can help Couples with Dementia Manage Reality Disjunction. In Proceedings of the 6th ACM Conference on Conversational User Interfaces (pp. 1-15). https://doi.org/10.1145/3640794.3665547

8.  Bennett, C. C. (2024). Findings from Studies on English-Based Conversational AI Agents (including ChatGPT) Are Not Universal. In *Proceedings of the 6th ACM Conference on Conversational User Interfaces (CUI)*, (pp. 1-5).

9.  Irfan, B., Kuoppamäki, S.M., & Skantze, G. (2023). Between reality and delusion: challenges of applying large language models to companion robots for open-domain dialogues with older adults. *ResearchSquare PrePrint*. https://doi.org/10.21203/rs.3.rs-2884789/v1

10. Banovic, S., Zunic, L. J., & Sinanovic, O. (2018). Communication difficulties as a result of dementia. *Materia socio-medica*, 30(3), 221.

11. Geraudie, A., Battista, P., García, A. M., Allen, I. E., Miller, Z. A., Gorno-Tempini, M. L., & Montembeault, M. (2021). Speech and language impairments in behavioral variant frontotemporal dementia: a systematic review. *Neuroscience & Biobehavioral Reviews*, 131, 1076-1095.

12. Robin, J., Harrison, J. E., Kaufman, L. D., Rudzicz, F., Simpson, W., & Yancheva, M. (2020). Evaluation of speech-based digital biomarkers: review and recommendations. *Digital Biomarkers*, 4(3), 99-108.

13. Hajjar, I., Okafor, M., Choi, J. D., Moore, E., Abrol, A., Calhoun, V. D., & Goldstein, F. C. (2023). Development of digital voice biomarkers and associations with cognition, cerebrospinal biomarkers, and neural representation in early Alzheimer's disease. *Alzheimer's & Dementia: Diagnosis, Assessment & Disease Monitoring*, 15(1), e12393.

14. Lin, H., Karjadi, C., Ang, T. F., Prajakta, J., McManus, C., Alhanai, T. W., ... & Au, R. (2020). Identification of digital voice biomarkers for cognitive health. *Exploration of Medicine*, 1, 406.

15. Yu, C., Sommerlad, A., Sakure, L., & Livingston, G. (2022). Socially assistive robots for people with dementia: Systematic review and meta-analysis of feasibility, acceptability and the effect on cognition, neuropsychiatric symptoms and quality of life. *Ageing Research Reviews*, 78, 101633. https://doi.org/10.1016/j.arr.2022.101633

16. Bennett, C. C., Kim, S. Y., Weiss, B., Bae, Y. H., Yoon, J. H., Chae, Y., ... & Shin, Y. (2024). Cognitive Shifts in Bilingual Speakers Affect Speech Interactions with Artificial Agents. *International Journal of Human–Computer Interaction*, 40(22), 7100-7111.

17. Forbes KE, Venneri A, & Shanks MF. (2002) Distinct patterns of spontaneous speech deterioration: an early predictor of Alzheimer's disease. *Brain and Cognition,* 48(2-3):356-361. PMID: 12030467.

18. Reilly, J., Rodriguez, A. D., Lamy, M., & Neils-Strunjas, J. (2010). Cognition, language, and clinical pathological features of non-Alzheimer's dementias: An overview. *Journal of Communication Disorders*, *43*(5), 438–452. https://doi.org/10.1016/j.jcomdis.2010.04.011

19. Forbes-McKay, K.E., & Venneri, A. (2005). Detecting subtle spontaneous language decline in early Alzheimer's disease with a picture description task. *Neurol Sci* 26, 243–254. https://doi.org/10.1007/s10072-005-0467-9

20. Ahmed, S., Haigh, A. M. F., de Jager, C. A., & Garrard, P. (2013). Connected speech as a marker of disease progression in autopsy-proven Alzheimer's disease. *Brain*, 136(12), 3727-3737. https://doi.org/10.1093/brain/awt269

21. Butters N, Delis DC, Lucas JA. (1995) Clinical assessment of memory disorders in amnesia and dementia. *Annu Rev Psychol*. 46:493-523. doi: 10.1146/annurev.ps.46.020195.002425. PMID: 7872736.

22. Calzà, L., Gagliardi, G., Favretti, R. R., & Tamburini, F. (2020). Linguistic features and automatic classifiers for identifying mild cognitive impairment and dementia. *Computer Speech & Language*, 65, 101113. https://doi.org/10.1016/j.csl.2020.101113

23. Fraser, K. C., Meltzer, J. A., & Rudzicz, F. (2015). Linguistic features identify Alzheimer's disease in narrative speech. *Journal of Alzheimer S Disease*, 49(2), 407–422. https://doi.org/10.3233/jad-150520

24. Luz, S., La Fuente Sofia, D., & Albert, P. (2018). A method for analysis of patient speech in Dialogue for Dementia Detection. *arXiv preprint*. https://doi.org/10.48550/arxiv.1811.09919

25. Jarrold, W., Nuance Communications, Peintner, B., Soshoma, Wilkins, D., Language & Linguistic Consulting, Vergryi, D., Richey, C., SRI International, Gorno-Tempini, M. L., Ogar, J., & University of California, San Francisco. (2014). Aided Diagnosis of Dementia Type through Computer-Based Analysis of Spontaneous Speech. In *Workshop on Computational Linguistics and Clinical Psychology: From Linguistic Signal to Clinical Reality* (pp. 27–37). https://aclanthology.org/W14-3204.pdf

26. Luz, S., Haider, F., de la Fuente Garcia, S., Fromm, D., & MacWhinney, B. (2021). Alzheimer's dementia recognition through spontaneous speech. *Frontiers in Computer Science*, 3, 780169. https://doi.org/10.3389/fcomp.2021.780169

27. Karlekar, S., Niu, T., & Bansal, M. (2018). Detecting linguistic characteristics of Alzheimer's dementia by interpreting neural models. *arXiv preprint*. https://doi.org/10.48550/arxiv.1804.06440

28. Orimaye, S. O., Wong, J. S., Golden, K. J., Wong, C. P., & Soyiri, I. N. (2017). Predicting probable Alzheimer's disease using linguistic deficits and biomarkers. *BMC Bioinformatics*, 18, 1-13.. https://doi.org/10.1186/s12859-016-1456-0

29. Glosser, G., & Deser, T. (1992). A comparison of changes in macrolinguistic and microlinguistic aspects of discourse production in normal aging. *Journal of Gerontology*, 47(4), P266-P272. https://doi.org/10.1093/geronj/47.4.P266

30. Ellis, D. G. (1996). Coherence patterns in Alzheimer's discourse. *Communication Research*, *23*(4), 472–495. https://doi.org/10.1177/009365096023004007

31. Dijkstra, K., Bourgeois, M. S., Allen, R. S., & Burgio, L. D. (2003). Conversational coherence: discourse analysis of older adults with and without dementia. *Journal of Neurolinguistics*, *17*(4), 263–283. https://doi.org/10.1016/s0911-6044(03)00048-4

32. Hoffman, P., Loginova, E., & Russell, A. (2018). Poor coherence in older people's speech is explained by impaired semantic and executive processes. *elife*, 7, e38907. https://doi.org/10.7554/eLife.38907

33. Hodges, J. R., Patterson, K., Oxbury, S., & Funnell, E. (1992). SEMANTIC DEMENTIA. *Brain*, *115*(6), 1783–1806. https://doi.org/10.1093/brain/115.6.1783.

34. Connor, L. T., & Obler, L. K. (2002). Anomia. In *Elsevier eBooks* (pp. 137–143). https://doi.org/10.1016/b0-12-227210-2/00027-3

35. Kirshner, H. S., Casey, P. F., Kelly, M. P., & Webb, W. G. (1987). Anomia in cerebral diseases. *Neuropsychologia*, *25*(4), 701–705. https://doi.org/10.1016/0028-3932(87)90062-5

36. Ralph, M. L. (2000). Classical anomia: a neuropsychological perspective on speech production. *Neuropsychologia*, *38*(2), 186–202. https://doi.org/10.1016/s0028-3932(99)00056-1

37. Woollams, A. M., Cooper-Pye, E., Hodges, J. R., & Patterson, K. (2008). Anomia: A doubly typical signature of semantic dementia. *Neuropsychologia*, *46*(10), 2503–2514. https://doi.org/10.1016/j.neuropsychologia.2008.04.005

38. Sheikh-Bahaei, N., Sajjadi, S. A., & Pierce, A. L. (2017). Current role for biomarkers in clinical diagnosis of Alzheimer disease and frontotemporal dementia. *Current Treatment Options in Neurology*, *19*(12). https://doi.org/10.1007/s11940-017-0484-z

39. Antonioni, A., Raho, E. M., Lopriore, P., Pace, A. P., Latino, R. R., Assogna, M., Mancuso, M., Gragnaniello, D., Granieri, E., Pugliatti, M., Di Lorenzo, F., & Koch, G. (2023). Frontotemporal dementia, where do we stand? A narrative review. *International Journal of Molecular Sciences*, *24*(14), 11732. https://doi.org/10.3390/ijms241411732

40. Gayraud, F., Lee, H., & Barkat-Defradas, M. (2010). Syntactic and lexical context of pauses and hesitations in the discourse of Alzheimer patients and healthy elderly subjects.

*Clinical Linguistics & Phonetics*, *25*(3), 198–209.
https://doi.org/10.3109/02699206.2010.521612

41. Klimova, B., & Kuca, K. (2016). Speech and language impairments in dementia. Journal of Applied Biomedicine, 14(2), 97-103. https://doi.org/10.1016/j.jab.2016.02.002

42. Szatloczki, G., Hoffmann, I., Vincze, V., Kalman, J., & Pakaski, M. (2015). Speaking in Alzheimer's Disease, is That an Early Sign? Importance of Changes in Language Abilities in Alzheimer's Disease. *Frontiers in Aging Neuroscience*, *7*. https://doi.org/10.3389/fnagi.2015.00195

43. Taler, V., Klepousniotou, E., & Phillips, N. A. (2009). Comprehension of lexical ambiguity in healthy aging, mild cognitive impairment, and mild Alzheimer's disease. *Neuropsychologia*, *47*(5), 1332–1343.

44. Zingeser, L. B., & Berndt, R. S. (1990). Retrieval of nouns and verbs in agrammatism and anomia. *Brain and Language*, *39*(1), 14–32. https://doi.org/10.1016/0093-934x(90)90002-x

45. Williams, E., Theys, C., & McAuliffe, M. (2023). Lexical-semantic properties of verbs and nouns used in conversation by people with Alzheimer's disease. *PLoS ONE*, *18*(8), e0288556. https://doi.org/10.1371/journal.pone.0288556

46. Tsanas, A., et al. (2012). Novel Speech Signal Processing Algorithms for High-Accuracy Classification of Parkinson's Disease. *IEEE Transactions on Biomedical Engineering, 59*(5), 1264–1271. https://doi.org/10.1109/tbme.2012.2183367

47. Rezaei, N., & Salehi, A. (2023). Acoustic Analysis of Speech. In *The Handbook of Clinical Linguistics* (pp. 1-12). University of Social Welfare and Rehabilitation Sciences.

48. Haider, F., De La Fuente, S., & Luz, S. (2019). An assessment of paralinguistic acoustic features for detection of Alzheimer's dementia in spontaneous speech. IEEE Journal of Selected Topics in Signal Processing, 14(2), 272-281. https://doi.org/10.1109/JSTSP.2019.2955022

49. Schuller, B., Steidl, S., & Batliner, A. (2009). The INTERSPEECH 2009 Emotion Challenge. *Interspeech 2009*. https://doi.org/10.21437/interspeech.2009-103

50. Mayle, A., Mou, Z., Bunescu, R. C., Mirshekarian, S., Xu, L., & Liu, C. (2019, September). Diagnosing Dysarthria with Long Short-Term Memory Networks. *Interspeech* (pp. 4514-4518). https://doi.org/10.21437/interspeech.2019-2903

51. Hernandez, A., Kim, S., & Chung, M. (2020). Prosody-based measures for automatic severity assessment of dysarthric speech. *Applied Sciences*, 10(19), 6999. https://doi.org/10.3390/app10196999

52. Nonavinakere Prabhakera, N., & Alku, P. (2018). Dysarthric Speech Classification Using Glottal Features Computed from Non-Words, Words and Sentences. *Aaltodoc (Aalto University)*. https://doi.org/10.21437/interspeech.2018-1059

53. Millet, J., & Zeghidour, N. (2022). Learning to Detect Dysarthria from Raw Speech. *ICASSP 2022 - 2022 IEEE International Conference on Acoustics, Speech and Signal Processing (ICASSP)* (pp. 5831–5835). https://doi.org/10.1109/icassp.2019.8682324

54. Skantze, G. (2021). Turn-taking in Conversational Systems and Human-Robot Interaction: A Review. *Computer Speech & Language, 67*, 101178. https://doi.org/10.1016/j.csl.2020.101178

55. Riest C., Jorschick A. B., & de Ruiter J. P. (2015). Anticipation in turn-taking: mechanisms and information sources. *Frontiers in Psychology,* 6:89. https://doi.org/10.3389/fpsyg.2015.00089

56. Levinson, S. C., & Torreira, F. (2015). Timing in turn-taking and its implications for processing models of language. *Frontiers in Psychology*, 6, 731. https://doi.org/10.3389/fpsyg.2015.00731

57. Heldner M. & Edlund J. (2010). Pauses, gaps and overlaps in conversations. *Journal of Phonetics,* 38 555–568 https://doi.org/10.1016/j.wocn.2010.08.002

58. Mueller K. D., Hermann B., Mecollari J., & Turkstra L. S. (2018). Connected speech and language in mild cognitive impairment and Alzheimer's disease: a review of picture description tasks. *J. Clin. Exp. Neuropsychol*, 40, 917–939. https://doi.org/10.1080/13803395.2018.1446513

59. Carlomagno, S., Santoro, A., Menditti, A., Pandolfi, M., & Marini, A. (2005). Referential communication in Alzheimer's type dementia. *Cortex, 41*(4), 520–534. https://doi.org/10.1016/S0010-9452(08)70192-8.

60. Pistono, A., et al. (2015). Pauses during autobiographical discourse reflect episodic memory processes in early Alzheimer's disease. *J. Alzheimers Dis, 50*(3), 687–698. https://doi.org/10.3233/JAD-150408.

61. De Looze, C., Dehsarvi, A., Crosby, L., Vourdanou, A., Coen, R. F., Lawlor, B. A., & Reilly, R. B. (2021). Cognitive and structural correlates of conversational speech timing in mild cognitive impairment and mild-to-moderate Alzheimer's disease: Relevance for early detection approaches. *Frontiers in Aging Neuroscience*, 13, Article 637404.

62. Chakraborty, R., Pandharipande, M., Bhat, C., & Kopparapu, S. K. (2020). Identification of dementia using audio biomarkers. *arXiv preprint*. https://doi.org/10.48550/arxiv.2002.12788

63. Rohanian, M., Hough, J., & Purver, M. (2020). Multi-Modal Fusion with Gating Using Audio, Lexical and Disfluency Features for Alzheimer's Dementia Recognition from Spontaneous Speech. *Interspeech 2022*. https://doi.org/10.21437/interspeech.2020-2721

64. Fraser, K. C., Fors, K. L., Eckerström, M., Öhman, F., & Kokkinakis, D. (2019). Predicting MCI status from multimodal language data using cascaded classifiers. *Frontiers in Aging Neuroscience*, 11. https://doi.org/10.3389/fnagi.2019.00205

65. Bullard, J., Alm, C. O., Liu, X., Yu, Q., & Proano, R. A. (2016, June). Towards early dementia detection: fusing linguistic and non-linguistic clinical data. In *Proceedings of* *the Third Workshop on Computational Linguistics and Clinical Psychology* (pp. 12-22). https://doi.org/10.18653/v1/W16-0302.

66. Russo, A., et al. (2019). Dialogue systems and conversational agents for patients with dementia: The human–robot interaction. Rejuvenation Research, 22(2), 109–120. https://doi.org/10.1089/rej.2018.2075.

67. Koutentakis, D., Pilozzi, A., & Huang, X. (2020). Designing socially assistive robots for Alzheimer's disease and related dementia patients and their caregivers: Where we are and where we are headed. Healthcare, 8(2). https://doi.org/10.3390/healthcare8020073

68. Zhou, D., Barakova, E. I., An, P., & Rauterberg, M. (2021). Assistant robot enhances the perceived communication quality of people with dementia: A proof of concept. IEEE Transactions on Human-Machine Systems, 52(3), 332–342. https://doi.org/10.1109/thms.2021.3112957.

69. Otaka, E., et al. (2024). Positive emotional responses to socially assistive robots in people with dementia: Pilot study. JMIR Aging, 7(1), e52443. https://doi.org/10.2196/52443

70. Whelan, S., Murphy, K., Barrett, E., Krusche, C., Santorelli, A., & Casey, D. (2018). Factors affecting the acceptability of social robots by older adults including people with dementia or cognitive impairment: A literature review. International Journal of Social Robotics, 10(5), 643–668. https://doi.org/10.1007/s12369-018-0471-x

71. Law, T., & Scheutz, M. (2021). Trust: Recent concepts and evaluations in human-robot interaction. *Trust in Human-Robot Interaction*, 27-57. https://doi.org/10.1016/B978-0-12-819472-0.00002-2

72. Koebel, K., Lacayo, M., Murali, M., Tarnanas, I., & Çöltekin, A. (2022). Expert insights for designing conversational user interfaces as virtual assistants and companions for older adults with cognitive impairments. *Chatbot Research and Design, 13171*, 23–38. https://doi.org/10.1007/s12369-018-0488-1.

73. Moro, C., Lin, S., Nejat, G., & Mihailidis, A. (2018). Social robots and seniors: A comparative study on the influence of dynamic social features on human–robot interaction. *International Journal of Social Robotics, 11*(1), 5–24.

74. Microsoft. (2024). Phi-3-Mini-4K-Instruct [Large language model]. *Hugging Face*. https://huggingface.co/microsoft/Phi-3-mini-4k-instruct

75. Biørn-Hansen, A., Majchrzak, T. A., & Grønli, T. M. (2017, April). Progressive web apps: The possible web-native unifier for mobile development. In *International Conference on Web Information Systems and Technologies,* (Vol. 2, pp. 344-351). https://doi.org/10.5220/0006353703440351.

76. Lanzi, A. M., Saylor, A. K., Fromm, D., Liu, H., MacWhinney, B., & Cohen, M. (2023). DementiaBank: Theoretical rationale, protocol, and illustrative analyses. American *Journal of Speech-Language Pathology*. https://doi.org/10.5220/0006353703440351/10.1044/2022_AJSLP-22-00281

77. Hsu, L. J., Bays, J. K., Tsui, K. M., & Sabanovic, S. (2023). Co-designing social robots with people living with dementia: Fostering identity, connectedness, security, and autonomy. In *Proceedings of the 2023 ACM Designing Interactive Systems Conference (DIS)*, pp. 2672-2688.

78. Hsu, L. J., Khoo, W., Swaminathan, M., Amon, K. J., Muralidharan, R., Satov, H., ... & Šabanović, S. (2024). Let's talk about you: Development and evaluation of an autonomous robot to support Ikigai reflection in older adults. In *33rd IEEE International Conference on Robot and Human Interactive Communication (ROMAN)*, pp. 1323-1330.

79. Bae, Y. H., & Bennett, C. C. (2023). Real-Time Multimodal Turn-taking Prediction to Enhance Cooperative Dialogue during Human-Agent Interaction. In *32nd IEEE International Conference on Robot and Human Interactive Communication (RO-MAN)* (pp. 2037-2044). https://doi.org/10.1109/ro-man57019.2023.10309569

80. Schuller, B., Steidl, S., Batliner, A., Hirschberg, J., Burgoon, J. K., Baird, A., Elkins, A., Zhang, Y., Coutinho, E., Evanini, K., et al. (2016). The interspeech 2016 computational paralinguistics challenge: Deception, sincerity & native language. *17th Annual Conference of the International Speech Communication Association (Interspeech)* (Vols 1-5), 2001–2005.

81. Tröger, J., Baykara, E., Zhao, J., Ter Huurne, D., Possemis, N., Mallick, E., ... & Ritchie, C. (2022). Validation of the remote automated ki: E speech biomarker for cognition in mild cognitive impairment: Verification and validation following DiME V3 framework. *Digital Biomarkers*, 6(3), 107-116.

82. Tombaugh, T. N., & McIntyre, N. J. (1992). The mini-mental state examination: a comprehensive review. *Journal of the American Geriatrics Society*, 40(9), 922-935.

83. Eyigoz, E., Mathur, S., Santamaria, M., Cecchi, G., Naylor, M., IBM Thomas J. Watson Research Center, & Pfizer Worldwide Research and Development. (2020). Linguistic markers predict onset of Alzheimer's disease. *EClinicalMedicine*, 28, 100583. https://doi.org/10.1016/j.eclinm.2020.100583

84. Hornecker, E., & Nicol, E. (2012). What do lab-based user studies tell us about in-the-wild behavior? Insights from a study of museum interactives. In *Proceedings of the 2023 ACM Designing Interactive Systems Conference (DIS)* (pp. 358-367).

85. Sabanovic, S., Michalowski, M. P., & Simmons, R. (2006). Robots in the wild: Observing human-robot social interaction outside the lab. In *9th IEEE International Workshop on Advanced Motion Control*, (pp. 596-601). IEEE.

86. Choi, S., Seo, J., Hernandez, M., & Kitsiou, S. (2024). Conversational agents in mHealth: use patterns, challenges, and design opportunities for individuals with visual impairments. *Journal of Technology in Behavioral Science*, 1-12.

87. McTear, M., Callejas, Z., Griol, D., McTear, M., Callejas, Z., & Griol, D. (2016). Conversational interfaces: devices, wearables, virtual agents, and robots. *The Conversational Interface: Talking to Smart Devices*, 283-308.

88. Hussain, T., Wang, D., & Li, B. (2024). The influence of the COVID-19 pandemic on the adoption and impact of AI ChatGPT: Challenges, applications, and ethical considerations. *Acta Psychologica*, 246, 104264.

89. Radhamani, R., Kumar, D., Nizar, N., Achuthan, K., Nair, B., & Diwakar, S. (2021). What virtual laboratory usage tells us about laboratory skill education pre-and post-COVID-19: Focus on usage, behavior, intention and adoption. *Education and Information Technologies*, 26(6), 7477-7495.

90. Bastarache, L., Brown, J. S., Cimino, J. J., Dorr, D. A., Embi, P. J., Payne, P. R., ... & Weiner, M. G. (2022). Developing real-world evidence from real-world data: Transforming raw data into analytical datasets. *Learning Health Systems*, 6(1), e10293.

91. Tang, A. S., Woldemariam, S. R., Miramontes, S., Norgeot, B., Oskotsky, T. T., & Sirota, M. (2024). Harnessing EHR data for health research. *Nature Medicine*, 30(7), 1847-1855.

92. Savitz, S. T., Savitz, L. A., Fleming, N. S., Shah, N. D., & Go, A. S. (2020, September). How much can we trust electronic health record data? *Healthcare*, 8(3), 100444. Elsevier Press.

93. Winkler, A., Kutschar, P., Pitzer, S., van der Zee-Neuen, A., Kerner, S., Osterbrink, J., & Krutter, S. (2023). Avatar and virtual agent-assisted telecare for patients in their homes: A scoping review. *Journal of Telemedicine and Telecare*, 1357633X231174484.

94. Abbott, M. B., & Shaw, P. (2016). Virtual Nursing Avatars: Nurse Roles and Evolving Concepts of Care. *Online Journal of Issues in Nursing*, 21(3), 7.

95. Sutton, R. T., Pincock, D., Baumgart, D. C., Sadowski, D. C., Fedorak, R. N., & Kroeker, K. I. (2020). An overview of clinical decision support systems: benefits, risks, and strategies for success. *NPJ Digital Medicine*, 3(1), 17.

96. Stojkovic, J., Choukse, E., Zhang, C., Goiri, I., & Torrellas, J. (2024). Towards Greener LLMs: Bringing Energy-Efficiency to the Forefront of LLM Inference. *arXiv preprint*. https://doi.org/10.48550/arXiv.2403.20306

97. Kim, Y., Xu, X., McDuff, D., Breazeal, C., & Park, H. W. (2024). Health-llm: Large language models for health prediction via wearable sensor data. *arXiv preprint*. https://doi.org/10.48550/arXiv.2401.06866

# Appendix

**Table A1:** Speech Biomarker Correlations with MMSE scores of PLwD Individuals (DementiaBank) – SEVERE CASES

|  | Grammar | Pragmatics | Anomia | Pronunciation | Prosody | Composite |
|---|---|---|---|---|---|---|
| **Min** | 0.2460 | 0.4870 | 0.2304 | 0.3750 | 0.3157 | 0.3989 |
| **Max** | 0.7895 | 0.8444 | 0.9230 | 0.5200 | 0.6443 | 0.6175 |
| **Avg** | 0.5377 | 0.6542 | 0.4438 | 0.4508 | 0.4331 | 0.5033 |
| **StdDev (avg)** | 0.1743 | 0.0825 | 0.1826 | 0.0094 | 0.0475 | 0.0533 |
| **Correlation w/ MMSE** | -0.3217 | -0.1944 | -0.2011 | 0.0715 | -0.1451 | -0.4327 |

**Table A2:** Speech Biomarker Correlations with MMSE scores of PLwD Individuals (DementiaBank) – MODERATE CASES

|  | Grammar | Pragmatics | Anomia | Pronunciation | Prosody | Composite |
|---|---|---|---|---|---|---|
| **Min** | 0.0924 | 0.3378 | 0.1788 | 0.3550 | 0.2838 | 0.3781 |
| **Max** | 0.9218 | 0.8208 | 0.8330 | 0.5350 | 0.6461 | 0.6162 |
| **Avg** | 0.5397 | 0.5985 | 0.4340 | 0.4464 | 0.4406 | 0.4906 |
| **StdDev (avg)** | 0.1837 | 0.0967 | 0.1176 | 0.0091 | 0.0529 | 0.0507 |
| **Correlation w/ MMSE** | -0.2921 | -0.1534 | -0.0540 | -0.1036 | -0.0663 | -0.3491 |

**Table A3:** Speech Biomarker Correlations with MMSE scores of PLwD Individuals (DementiaBank) – MILD CASES

|  | Grammar | Pragmatics | Anomia | Pronunciation | Prosody | Composite |
|---|---|---|---|---|---|---|
| **Min** | 0.0000 | 0.4209 | 0.2208 | 0.3600 | 0.3035 | 0.3726 |
| **Max** | 0.8214 | 0.7648 | 0.7211 | 0.5400 | 0.6441 | 0.6413 |
| **Avg** | 0.4434 | 0.5847 | 0.4210 | 0.4409 | 0.4533 | 0.4688 |
| **StdDev (avg)** | 0.1922 | 0.0836 | 0.1157 | 0.0133 | 0.0619 | 0.0644 |
| **Correlation w/ MMSE** | -0.2118 | -0.0394 | -0.0316 | 0.1149 | -0.0002 | -0.2081 |

**Table A4:** Speech Biomarker Correlations with MMSE scores of PLwD Individuals (DementiaBank) – CONTROL CASES (no dementia)

|  | Grammar | Pragmatics | Anomia | Pronunciation | Prosody | Composite |
|---|---|---|---|---|---|---|
| **Min** | 0.0000 | 0.3887 | 0.1330 | 0.3600 | 0.2723 | 0.2661 |
| **Max** | 0.7147 | 0.7392 | 0.8943 | 0.5400 | 0.6457 | 0.5455 |
| **Avg** | 0.3234 | 0.5442 | 0.3706 | 0.4446 | 0.4430 | 0.4212 |
| **StdDev (avg)** | 0.1730 | 0.0729 | 0.1006 | 0.0087 | 0.0531 | 0.0515 |
| **Correlation w/ MMSE** | -0.1230 | -0.1665 | 0.0710 | 0.0077 | -0.0287 | -0.1121 |